%% file: main.tex
\documentclass[10pt,twocolumn,letterpaper]{article}

\usepackage[camera-ready]{iccv}      % To produce the REVIEW version
\usepackage{multirow}
\usepackage{wasysym}
\usepackage{colortbl}
\usepackage{marvosym}

\newcommand{\mo}[1]{#1}

\definecolor{iccvblue}{rgb}{0.21,0.49,0.74}
\usepackage[pagebackref,breaklinks,colorlinks,allcolors=iccvblue]{hyperref}

\def\paperID{5483} % *** Enter the Paper ID here
\def\confName{ICCV}
\def\confYear{2025}

\title{Progressive Homeostatic and Plastic Prompt Tuning for Audio-Visual Multi-Task Incremental Learning}

\author{Jiong Yin$^{1,2}$\thanks{This work is done during the intern in VIPL group, ICT, CAS.} $\phantom{00}$ Liang Li$^{2}$ \thanks{Corresponding author} $\phantom{00}$ Jiehua Zhang$^3$ $\phantom{00}$ Yuhan Gao$^1$ $\phantom{00}$ Chenggang Yan$^1$ $\phantom{00}$ Xichun Sheng$^4$ $\phantom{00}$\\
{\small $^1$Hangzhou Dianzi University~ $^2$Institute of Computing Technology, Chinese Academy of Sciences} \\
{\small $^3$Xi'an Jiaotong University~ $^4$Macao Polytechnic University} \\
{\tt\scriptsize\{jiong.yin,yuhangao,cgyan\}@hdu.edu.cn $\phantom{00}$ liang.li@ict.ac.cn $\phantom{00}$ jiehua.zhang@stu.xjtu.edu.cn $\phantom{00}$ p2314922@mpu.edu.mo}}

\begin{document}
\maketitle
\input{sec/0_abstract}

\input{sec/1_introduction}
\input{sec/2_related_work}

\input{sec/3_method}

\input{sec/4_experiments}
\input{sec/5_conclusion}
\input{sec/6_ackownledgement.tex}
% \input{sec/X_suppl}
% \clearpage
{
    \small
    \bibliographystyle{ieeenat_fullname}
    \bibliography{main}
}
\clearpage
\input{sec/X_suppl}
\end{document}

%% file: sec/0_abstract.tex
\begin{abstract}

Audio-visual multi-task incremental learning aims to continuously learn from multiple audio-visual tasks without the need for joint training on all tasks.
% However, it presents challenges in balancing task interference and knowledge sharing, as well as preserving modality specificity and integrating modality generality.
% 
The challenge of the problem is how to preserve the old task knowledge while facilitating the learning of new task with previous experiences. 
To address these challenges, we introduce a three-stage Progressive Homeostatic and Plastic audio-visual prompt (PHP) method.
% that retains task-specific prompts while adapting shared parameters for new tasks.
% we design
In the shallow phase, we design the task-shared modality aggregating adapter to foster cross-task and cross-modal audio-visual representation learning to enhance shared understanding between tasks.
In the middle phase, we propose the task-specific modality-shared dynamic generating adapter, which constructs prompts that are tailored to individual tasks while remaining general across modalities, which balances the model’s ability to retain knowledge against forgetting with its potential for versatile multi-task transferability.
In the deep phase, we introduce the task-specific modality-independent prompts to further refine the understand ability by targeting individual information for each task and modality.
% By incorporating these three phases, PHP aims to enhance the model’s adaptability to changing task scenarios and improve its performance across various audio-visual tasks. This comprehensive approach demonstrates its effectiveness in addressing the challenges of audio-visual multi-task incremental learning.
By incorporating these three phases, PHP retains task-specific prompts while adapting shared parameters for new tasks to effectively balance knowledge sharing and specificity. 
Our method achieves SOTA performance in different orders of four tasks~(AVE, AVVP, AVS and AVQA). Our code can be available at \href{https://github.com/ENJOY-Yin-jiong/PHP}{https://github.com/ENJOY-Yin-jiong/PHP}.

\end{abstract}

%% file: sec/1_introduction.tex
\section{Introduction}
\begin{figure}[tp]
    \centering
    % \vspace{-1.5em}
    % \setlength{\abovecaptionskip}{0.01cm}
    % \setlength{\belowcaptionskip}{0.01cm}
    \includegraphics[scale=0.42]{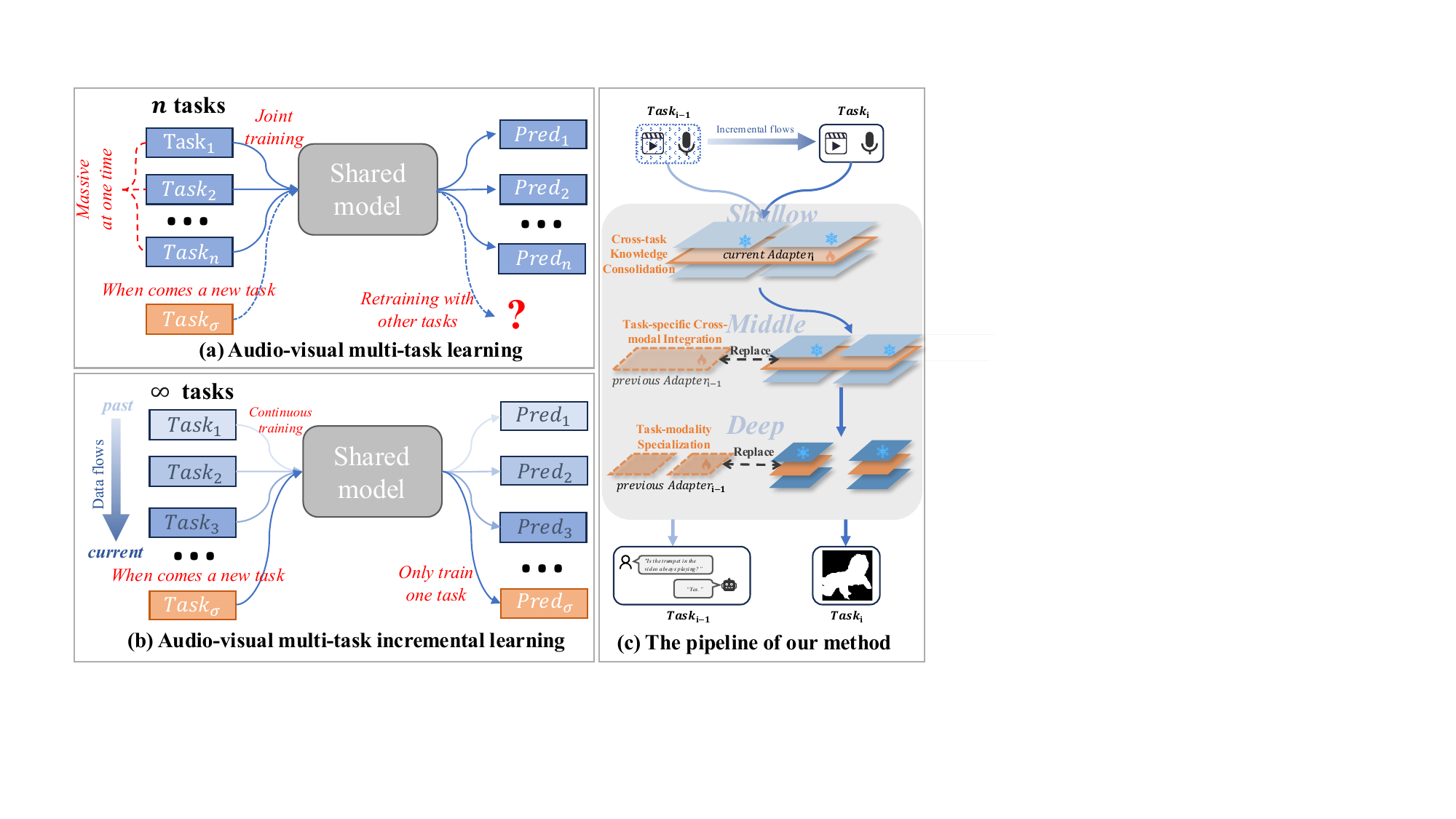} % 0.63  0.67
    \caption{
    (a) \textbf{Audio-visual multi-task learning} allows simultaneous training on various audio and video tasks. But, the model may require retraining when a new task emerges.
    % 在能在保持对前序任务不忘记的同时，以更灵活的方式应对信道来的任务
    (b) \textbf{Audio-visual multi-task incremental learning} is able to address new tasks more flexibly while maintaining the memory of previous tasks.
    (c) \textbf{The pipeline of our method.}
    }
    \label{fig_intro}
    \vspace{-1.5em}
\end{figure}

% 视觉和听觉是人类感知和理解世界的基础。
Visual and auditory perception form the core of human understanding of the world.
Audio-visual multi-task learning attracts significant research interest for integrating audio-visual cues to address various tasks.
% 然而，由于统一训练的问题，多任务模型不具备足够的灵活性以应对不同的任务。
However, due to the dynamic nature of the real world, once a multi-task model trained uniformly is deployed, it often struggles to update continuously to acquire new knowledge, while also fulfilling new task requirements as illustrated in Fig~\ref{fig_intro}~(a).

% % 第二段
% % audio-visual multi-task incremental learning旨在在音视频多任务数据流中进行持续学习，而无需对多个多模态任务进行同时训练。
% One way to adapt dynamic data flows is incremental learning.
% Audio-visual multi-task incremental learning aims to engage in continuous learning within audio-visual multi-task data streams without the need for simultaneous training on multiple multi-modal tasks.
% % 增量学习策略能帮助模型极大增强多任务模型的灵活性，能够根据共享知识快速适应新任务。
% The incremental learning strategy can greatly enhance the flexibility of multi-task models and enable them to quickly adapt to new tasks based on shared knowledge, as shown in Fig~\ref{fig_intro}~(b).
% % 增量学习致力于缓解模型灾难性遗忘问题，最近的一些工作已经取得了较大的突破。 两篇文章
% Recent works in incremental learning are dedicated to alleviating the catastrophic forgetting problem~\cite{nguyen2019toward} and have made significant breakthroughs. 
% % 一类方法是通过调整和优化模型结构或参数来适应新数据，
% One part of works focuses on involves adjusting and optimizing the model’s architecture or parameters to adapt to new data. However, these methods present potential issues regarding data security and the need for substantial data storage capabilities.
% As another part of the incremental learning topic, prompt-based methods~\cite{sprompt, dualprompt, l2p, pc} avoid the aforementioned issues by leveraging context-rich prompts to guide the model’s adaptation to new data without the need for extensive retraining or parameter adjustments.
% % prompt总结。

Audio-visual multi-task incremental learning aims to conduct continuous learning within audio-visual task streams without the need for simultaneous training on all tasks. 
% As illustrated in Fig~\ref{fig_intro}~(a), conventional approaches typically require joint training with complete access to all task data, which is computationally intensive and impractical for real-world deployments. In contrast, 
An incremental learning strategy can greatly enhance the flexibility of multi-task models and enable them to quickly adapt to new tasks based on shared knowledge, as shown in Fig~\ref{fig_intro}~(b).
\mo{However, this approach faces the critical challenge of catastrophic forgetting, where new knowledge overwrites existing representations, causing significant performance decline on previously learned tasks.}
% Audio-visual multi-task incremental learning enables continuous learning without simultaneous training on all tasks, enhancing model flexibility and adaptation to new tasks using shared knowledge. However, this approach faces catastrophic forgetting, where new knowledge overwrites existing representations, degrading performance on previously learned tasks.

Previous works in incremental learning have made significant progress in mitigating catastrophic forgetting. Some methods maintain memory of old tasks through parameter regularization~\cite{aljundi2018memory, kirkpatrick2017overcoming, li2017learning, zenke2017continual}, while others rely on data replay~\cite{buzzega2020dark, cha2021co2l, chaudhry2019tiny, wu2019large, chen2023promptfusion, ebrahimi2020adversarial, pham2021dualnet, prabhu2020gdumb} to revisit past knowledge. Recently, prompt-based methods~\cite{huang2022unsupervised, jung2023generating, smith2023coda, sprompt, dualprompt, l2p, pc} show promising potential by introducing a small number of trainable prompt parameters to adapt to new tasks, avoiding large-scale parameter updates and data storage. 
% 忽视了音视频特有的挑战
\mo{However, these methods primarily focus on single-modality scenarios and do not fully address audio-visual specific challenges.}

% 第三段
% 然而，audio-visual multi-task incremental learning带来了全新的挑战：
% However, audio-visual multi-task incremental learning brings about brand new challenges:
Extending incremental learning to audio-visual multi-task scenarios introduces fundamental challenges beyond traditional single-modality settings:
% 1）任务间干扰和知识共享间的平衡。 不同任务之间共享通用性特征或知识可以促进单一任务的学习效果，但同时这种共享也可能导致任务间的相互干扰。 同时，由于任务间的差异，过多的共享知识又会导致增量过程不可避免地带来灾难性遗忘问题。 
1) \textit{Balancing interference and knowledge sharing between tasks.} 
% Sharing common knowledge across different tasks can enhance the learning effect for individual tasks, but this sharing may also lead to mutual interference between tasks. Moreover, due to the differences between tasks, excessive sharing of knowledge can result in catastrophic forgetting issues that are inevitable in the incremental process.、
Sharing knowledge across tasks boosts individual learning but may cause interference between different tasks. Additionally, over-sharing due to task differences may exacerbate the catastrophic forgetting issue during incremental learning.
% 2） 模态特异性保留与模态通用性融合的平衡。 在视听多任务学习过程中，掌握视听通用性表示有助于模型发现不同模态之间的内在联系，从而提升模型的泛化能力。 然而，过度融合音视频模态可能会导致单一模态的原始信息丢失，这可能会干扰模型在特定任务中对单一模态关键信息的保留。
% During audio-visual multi-task learning, the model benefits from mastering general representations across audio and visual modalities, which reveals intrinsic connections and improves its generalization abilities. Nevertheless, an excessive fusion of these modalities risks losing raw information from each, potentially hindering the model’s ability to retain crucial details from one modality in certain tasks. This loss of individual modality information can disrupt the model’s performance on specific tasks that rely on single-modality insights.
2) \textit{Balancing the preservation of modality specificity and the integration of modality generality.} 
During audio-visual multi-task learning, the model benefits from mastering general representations across audio and visual modalities, which reveals intrinsic connections and improves its transfer ability. 
% Nevertheless, excessive fusion risks losing raw information from each modality, potentially hindering the model's ability to retain crucial details from one modality for specific tasks. This loss of individual modality information can disrupt the model’s performance on specific tasks that rely on single-modality insights.
\mo{However, there exists a fundamental trade-off between cross-modal fusion and modality-specific information preservation, which can impact performance on tasks requiring detailed single-modality features.
% 举例，ave需要对每一模态都进行定位。
For example, the AVE task requires precise localization within each modality separately.}

% 第四段
% 为什么要加上
% 为此， 我们提出了a novel Progressive homeostatic plastic audio-visual prompt方法， 旨在通过三个阶段策略来有效缓解增量遗忘问题，并显著提升知识共享效率。
% To address this, we propose a novel Progressive Homeostatic and Plastic audio-visual prompt~(PHP) method, designed to effectively mitigate the issue of incremental forgetting through a three-phase strategy, while significantly enhancing the capability of knowledge sharing.
% 受。。。的启发
\mo{Inspired by the knowledge consolidation pathway observed in continual pre-trained models, where semantic circuits evolve from deep conceptual abstraction to shallow representation refinement~\cite{ou2025llms},}
we introduce a novel Progressive Homeostatic and Plastic audio-visual prompt~(PHP) method to tackle incremental forgetting issue and boost knowledge sharing with a three-stage manner.
\mo{
The proposed framework operates through three hierarchically organized phases: (a) cross-task knowledge consolidation at shallow layers, which transitions into (b) task-specific cross-modal integration in middle layers, ultimately culminating in (c) task-modality specialization at deep layers.}
% 在浅层阶段，我们提出了Task-shared modality aggregating Adapter， 其通过跨任务和跨模态的深度融合，使模型能够学习到更为通用的音视频表示，从而最大化不同任务间的知识共享。
% In the shallow phase, we propose the task-shared modality aggregating adapter, which, through deep fusion across tasks and modalities, enables the model to learn more generalized audio-visual representations, thereby maximizing knowledge sharing between different tasks.
In the shallow phase, the Task-shared Modality Aggregating~(TMA) adapter 
% that fuses information comprehensively across tasks and modalities to learn more generalized audio-visual representations.
\mo{creates the universal audio-visual representations through various cross-modal fusions.}
% This primary fusion maximizes knowledge sharing between various tasks to enhance the fundamental audio-visual correspondence.
\mo{This foundational stage maximizes knowledge sharing by establishing task-agnostic correspondences critical for subsequent adaptations.}
% 在中层阶段，我们提出了Task-specific modality-shared dynamic generating adapter，其通过大范围的prompt pool来充分学习视听表征以增强模型视听感知能力，随后动态选择出实例级prompt，以学习特定任务的音视频融合上下文。
In the middle stage, the Task-specific Modality-shared Dynamic Generating (TMDG) adapter enhances audio-visual representations using a prompt pool to improve perception. It then selects instance-level prompts to customize audio-video fusion for each task's specific needs.
% The middle-phase​ Task-specific Modality-shared Dynamic Generating (TMDG) adapter ​addresses task interference​ via ​two innovations: ​1) a multimodal prompt pool​ capturing diverse audio-visual relationships, and ​2) dynamic instance-level selection​ that ​tailors fusion strategies​ to specific task requirements.
% 在深层阶段，我们提出了Task-specific modality-independent prompts，通过任务以及模态间的隔离的prompt，以保持模态特异性的同时挖掘特定任务的深层次表征，以此增强模型抗遗忘能力。
At the deep stage, Task-specific Modality-Independent~(TMI) prompts
operate in isolation for each task and modality, ensuring the preservation of modality-specific details while excavating profound representations tailored to individual tasks. This reinforces the model’s resilience against the forgetting of information and strengthens its overall performance.
% At the deep stage, Task-specific Modality-Independent (TMI) prompts ​implement dual isolation: ​1) modality-specific pattern preservation​ (e.g., acoustic fingerprints in audio streams) and ​2) task-defining feature stabilization​ (e.g., scene context in video frames). ​This final defense layer​ prevents catastrophic forgetting by ​decoupling critical knowledge​ from transient task adaptations.
% 整体性表述
% \mo{Building upon the foundational cross-modal representations established by TMA, the model progressively develops task-aware adaptations through TMDG, which in turn creates the prerequisite for preserving modality-specific details via TMI.}

The main contributions of this paper as follows:
\begin{itemize}
    \item 
    % 通过引入增量策略，模型能够以更灵活的方式应对任务场景变化。
    % We propose the Progressive Homeostatic and Plastic audio-visual prompt~(PHP) method. By introducing an incremental strategy, the model can adapt to changes in task scenarios more flexibly.
    We propose a Progressive Homeostatic and Plastic (PHP) audio-visual prompting framework that enables incremental learning for dynamic audio-visual task streams.
    
    \item 
    % 我们设计TMA模块， 它能在模型的浅层种
    % We design the task-shared modality aggregating adapter, which is injected into the shallow phase of the pre-trained models to achieve the general correspondence between audio and visual in joint tasks.
    We design a task-shared modality aggregating adapter in shallow layers that learns universal audio-visual representations through complementary attention mechanisms.

    \item 
    % we devise the task-specific modality-shared dynamic generating adapter and task-specific modality-independent prompts in the middle and deep phases to explore the clue tailored to the specific task while mitigating catastrophic forgetting issues.
    We introduce a hierarchical prompting strategy combining task-specific modality-shared and modality-independent prompts to balance knowledge preservation and task adaptation.

    \item Extensive experiments on four audio-visual tasks (AVE, AVVP, AVS and AVQA) demonstrate that our method achieves state-of-the-art performance in both knowledge preservation and transfer capability.
% Experiments on three audio-visual tasks demonstrate superior performance over existing methods in both knowledge preservation and transfer capability.
% efine audio-visual processing and maintain modal-specific intricacies for optimized task adaptation.
    
\end{itemize}

%% file: sec/2_related_work.tex
\section{Related Work}
% audio-visual representations learning
% 再写细一点
Deep learning’s development has propelled significant advancements in the field of computer vision~\cite{cym_1,gby_1,yan_1,yan_2,yan_4,yan_5,yan_6,yan_7,yan_8,yan_9,yan_10,zjk_1,li_1,Li_2,li_3,li_4} and multi-modal learning~\cite{zzd_1, zzd_2,zzq_1,yan_3,li_5,li_6}.
\subsection{Audio-visual Understanding}
% 旨在做什么
Audio-visual understanding tasks focus on learning the correspondence between audio and visual modalities from videos to achieve multi-modal perception.
% 引入多模态任务，一些工作
This has led to advancements in a range of audio-visual tasks, such as audio-visual event localization~\cite{ave,ave_1,ave_2,ave_3,ave_4}, audio-visual parsing~\cite{avvp,avvp_1,avvp_2,avvp_3,avvp_4}, audio-visual question answer~\cite{music-avqa,avqa_1,avqa_2,avqa_3}, audio-visual spatialization~\cite{avs, avs_1,avs_2}. 
% 我们的改进
In this work, we explore the transferrable representations progressively, incrementally engaging with various audio-visual tasks. 
% 遗忘的同时需要考虑泛化性
It is more challenging than the single task above as forgetting and transferability should be considered.

% incremental learning
\subsection{Incremental Learning}
Incremental learning aims to continuously integrate new knowledge without compromising the integrity of previously learned information, thereby maintaining a robust and adaptable representation over time. 
% 灾难性遗忘
Its major challenge is the catastrophic forgetting problem~\cite{mccloskey1989catastrophic}, which occurs when the model’s performance on previously learned tasks significantly degrades as it learns new tasks. 
% Rehearsal-based techniques maintain a subset of past data to revisit old knowledge, but they are sensitive to data selection and face issues with privacy and memory. Despite these approaches, the field still grapples with data privacy and memory constraints.
% 
To reduce the issue, 
% 这个方法都可以不要， 看篇幅
architecture-based methods often extract features from multiple layers~\cite{ke2020continual, mallya2018packnet, serra2018overcoming} or extend models~\cite{li2019learn, rajasegaran2019random, rusu2016progressive, shang2023incrementer, yan2021dynamically, yoon2017lifelong}, but the effectiveness of these models is severely limited due to the complexity of the extended models.
Regularization-based methods~\cite{aljundi2018memory, kirkpatrick2017overcoming, li2017learning, zenke2017continual} impose constraints on loss functions to preserve old knowledge, but their effectiveness depends on the relationship between tasks, limiting their use in complex scenarios.
Rehearsal-based methods maintain a subset of past data to revisit old knowledge~\cite{buzzega2020dark, cha2021co2l, chaudhry2019tiny, wu2019large, chen2023promptfusion, ebrahimi2020adversarial, pham2021dualnet, prabhu2020gdumb}, but they are sensitive to data selection~\cite{buzzega2020dark, cha2021co2l, chaudhry2019tiny, wu2019large, hu2022curiosity, yu2023contrastive} and face issues with privacy and memory~\cite{shokri2015privacy}. 
% Despite these approaches, the field still grapples with data privacy and memory constraints.

\begin{figure*}[tp]
    \centering
    \vspace{-1.5em}
    \setlength{\abovecaptionskip}{0.01cm}
    \setlength{\belowcaptionskip}{0.01cm}
    \includegraphics[scale=0.60]{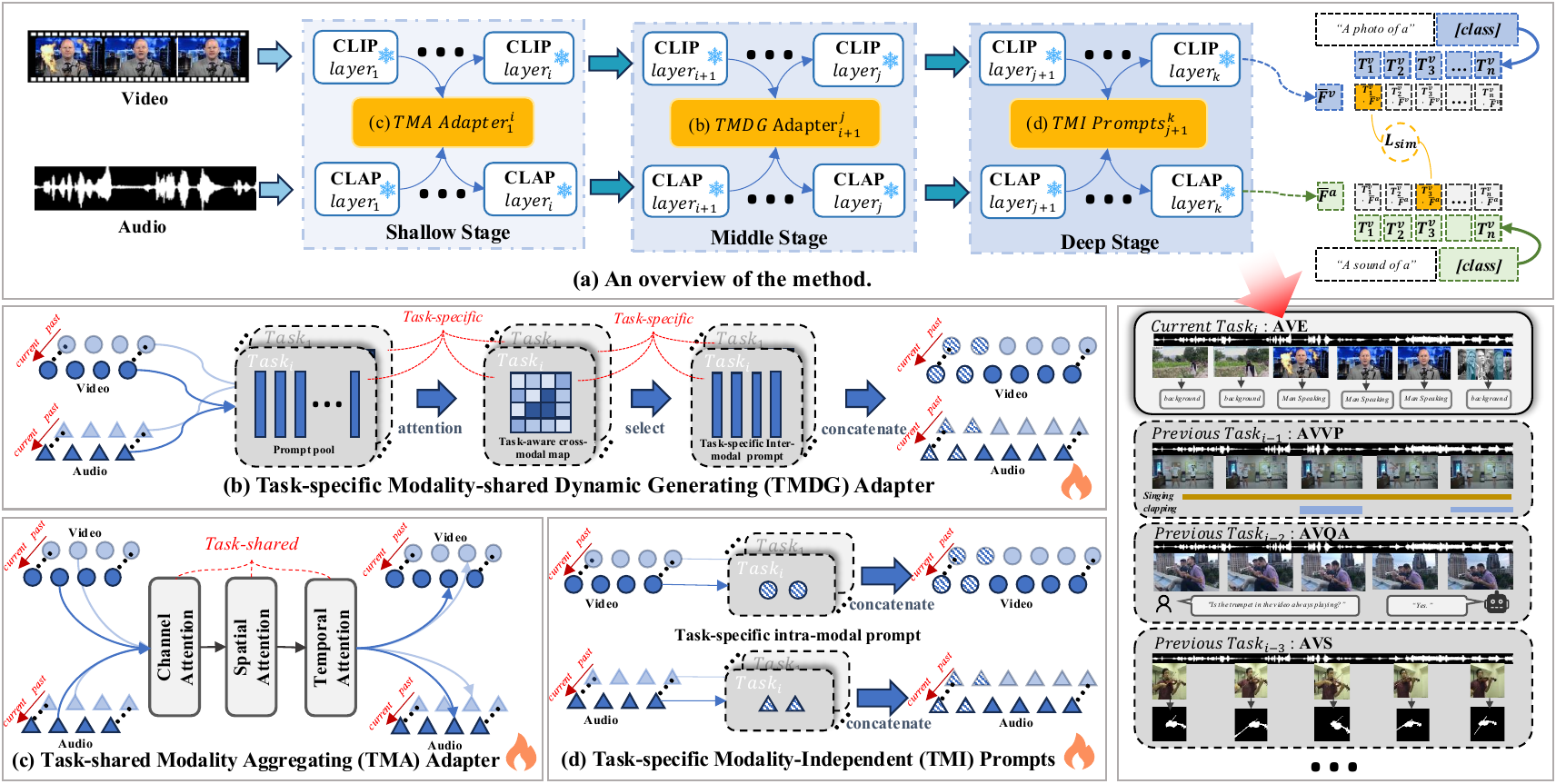} % 0.63  0.67
    \caption{
    The Progressive Homeostatic and Plastic audio-visual prompt (PHP) framework consists of three stages for balancing knowledge preservation and transfer: (a) an overview of the three-stage architecture, (b) a task-specific modality-shared dynamic generating adapter for balancing anti-forgetting and multi-task transfer abilities, (c) a task-shared modality aggregating adapter in shallow stage for learning cross-task and cross-modal representations, and (d) task-specific modality-independent prompts in deep stage for preserving task-specific and modality-specific features.
    }
    \label{fig2}
    % \vspace{-10pt}
    \vspace{-1.5em}
\end{figure*}

\subsection{Prompt Tuning for Incremental Learning}
Recent works~\cite{bahng2022visual, douillard2022dytox, huang2022unsupervised, jung2023generating, smith2023coda, sprompt, dualprompt, l2p, pc,gong_1,gong_2,gong_3,gong_4,dph_1,yzw_1} in incremental learning show that prompt tuning is an efficient method for adapting models by training only new inputs and reusing existing knowledge without full retraining.
L2P~\cite{l2p}, as the pioneer method of prompt tuning in incremental learning, introduces prompts as trainable parameters in continual learning, bypassing the need for rehearsal by selecting from a prompt pool.
DualPrompt~\cite{dualprompt} innovates by dividing the prompt pool into task-agnostic and task-specific subsets for more efficient learning.
Similar to DualPrompt, S-Prompt~\cite{sprompt} independently learns task-specific prompts to reduce catastrophic forgetting through separate tuning strategies. However, this leads to an increase in prompts with the number of tasks, with selection based on a one-hot indexing method.
As an improvement, PC~\cite{pc} generates more expressive prompt representations by creating instance-specific prompts from a predefined prompt codebook.
% 在这篇工作中，我们将通过三个阶段来进行音视频多任务增量学习。
In this work, we conduct audio-video multi-task incremental learning through three stages to simultaneously enhance the model’s anti-forgetting capability and multi-task transferability.

%% file: sec/3_method.tex
\section{Proposed Approach}

% 任务
\subsection{Overview}
% mmm: 与intro一致
% 1. 介绍视听任务
% audio-visual multi-task incremental learning 旨在在视听任务数据流中进行持续学习。以此让模型适应实时变化的数据以及不断到来的新任务。
% 这个任务的难度在于：1)   2)
% 在本工作中，我们将着重在 xxx 三个任务上进行多任务增量学习。
Audio-visual multi-task incremental learning aims to conduct continuous learning in the data stream of audio-visual tasks. This allows the model to adapt to changing data in real-time and continuously emerge new tasks.
The challenge lies in:
1) Balancing interference between tasks and knowledge sharing.
2) Balancing the preservation of modality specificity and the integration of modality generality.
% To address these challenges, we propose a Progressive Homeostatic and Plastic (PHP) framework that processes audio-visual information through three carefully designed stages.

% 2. 介绍方法的整体流程
% 在我们的方法中，
% 我们 以 CLIP 和CLAP作为backbone，以增量的方式来统一视听多任务学习。
Our framework utilizes CLIP~\cite{clip} and CLAP~\cite{clap} as the backbone with three complementary components operating at different network depths:
% 在shallow阶段，我们通过 Task-shared modality aggregating Adapter 来学习跨任务和跨模态的视听通用表征。
In the shallow stage, we employ a Task-shared Modality Aggregating (TMA) adapter to learn universal audio-visual representations that can benefit multiple tasks. This stage emphasizes the extraction of fundamental cross-modal correspondences, which serve as shared foundational features across diverse audio-visual tasks.
% 在middle阶段，Task-specific Modality-shared Dynamic Generating Adapter 被设计用来生成任务特定但模态通用的prompt，以此来兼顾模型的抗遗忘能力以及模型的多任务泛化性。
In the middle stage, The Task-specific Modality-shared Dynamic Generating (TMDG) adapter is designed to balance task specificity and modality sharing. This component generates adaptive prompts that help maintain task-specific knowledge while enabling cross-modal feature integration.
% 在deep阶段，Task-specific Modality-independent Prompts 以任务特定和模态特定的特性来引导模型学习特定信息。
In the deep stage, we develop Task-specific Modality-independent (TMI) prompts to preserve the unique characteristics of each modality and task. This ensures that critical task-specific and modality-specific information is retained during incremental learning.

\subsection{Problem Formulation}
% 对任务进行定义
% 给定
Assuming $\mathcal{D} = \left \{\mathcal{D}_1, \mathcal{D}_2, ..., \mathcal{D}_T \right \}$ as a sequence of tasks with $T$ incremental tasks, where \textit{t-th} audio-visual task $\mathcal{D}_t= \left \{(a^{t}_{i}, v^{t}_{i}, y^{t}_{i})\right \}^{N_t}_i$ contains tuples of audio-visual pair $a^{t}_{i} \in \mathcal{A}^t$ and $ v^{t}_{i} \in \mathcal{V}^t$ and the coresponding ground-truth labels $y^t_i \in \mathcal{Y}^t$. The audio-visual multi-task incremental learning is defined to train a single model $f_\theta:(\mathcal{A}, \mathcal{V}) \rightarrow \mathcal{Y} $ with $\theta$ parameters to handle the $T$ different audio-visual tasks. 
% 在训练阶段，模型可知当前在做的任务，以便模型更新并生成合适的输出。
During the training and inferring phase, the current task is available, which allows the model to update the suitable prediction module and generate appropriate outputs.

\subsection{Task-shared Modality Aggregating Adapter}
% 讲讲为什么要模态和任务共享。
% 模型在浅层主要关注保留数据的局部结构和增强鲁棒性，而深层学习模型则能够学习到更复杂的特征表示和揭示数据中隐藏的深层信息。
% 模型在浅层主要学习数据的局部结构和特征表示，而在深层则能够学习到更复杂的判别性信息和潜在概念以应对不同任务，并且可以更好地处理大规模数据和高复杂度任务。
Zhang \textit{et al.}~\cite{zhang2021survey} indicates that the shallow stage of a model learns the local structure and feature representation of the data. At the deeper stage of a model, it is capable of learning more complex discriminative information and latent concepts to address different tasks more effectively.
% 受其启发，我们首先提出了Task-shared Modality Aggregating Adapter 来让模型在浅层优先学习跨任务跨模态的音视频表征
Inspired by this, we first propose the Task-shared Modality Aggregating~(TMA) adapter to enable the model to prioritize learning cross-task, cross-modal audio-video representations at the shallow stage.

% 具体而言，TMA adapter 采用了
% 整体上，TMA 由channel attention, spatial attention 和 temporal attention 组成，以多角度的形式来促进音视频多模态融合。
Overall, drawing insights by DG-SCT~\cite{duan2024cross}, TMA adapter consists of channel attention, spatial attention, and temporal attention fusion, which together facilitate the fusion of audio-video multi-modal data from multiple perspectives.
% 具体的，在channel attention fusion 中，我们让音频和视频互为引导信息，以此从通道维度提取有价值信息，公式表示为：
% Specifically, in the channel attention fusion, we use audio and video as guiding information for each other to extract valuable information from the channel dimension, which can be represented by the following formula:
Specifically, channel attention enables bidirectional feature refinement between modalities by identifying semantically related channel patterns. 
% We use audio and video as guiding information for each other to extract valuable information from the channel dimension, which can be represented by the following formula:
\mo{For an input video feature $v^t$ and audio feature $a^t$, the audio-to-video channel attention map $M^t_{ac}$ is computed through global statistics aggregation and channel projection:}
% \begin{equation}
%     M^t_{vc} = \sigma(\Phi_a(\delta_a(a^t_c))) \in \mathbb{R}^{C_v\times1}
% \end{equation}
\begin{equation}
    M^t_{vc} = \sigma\left( W_v \cdot \delta_v \left( \Phi_a(a^t) \right) \right) \in \mathbb{R}^{C_v \times 1},
\end{equation}
\mo{where $\Phi_a(\cdot)$ denotes global average pooling over spatial-temporal audio dimensions, $\delta_v(\cdot)$ projects pooled features to video channels, and $W_v$ implements learnable spatial weighting. Conversely, the video-to-audio channel adaptation follows:}
\begin{equation}
    M^t_{ac} = \sigma\left( W_a \cdot \delta_a \left( \Phi_v(v^t) \right) \right) \in \mathbb{R}^{C_a \times 1},
\end{equation}
\mo{with $\Phi_v(\cdot)$ aggregating video spatial features and $\delta_a(\cdot)$ aligning dimensions. The refined features are then obtained as $v_{cm}^t=v^t \odot M^t_{vc} $ and $a^t_{cm} = a^t \odot M^t_{ac}$ , where $\odot$ denotes channel-wise multiplication.}
% where $M^t_{vc}$ and $M^t_{ac}$ are the A2V and V2A channel attention maps, respectively. $\sigma$ is a sigmoid function. $\Phi_a$ and $\Phi_v$ denote the spatial-wise global average pooling. $\delta_a$ and $\delta_v$ represent the projection layer for audio and visual, respectively. 
% $a^t_{cm} = a^t \odot v^t$ and $a^t_{cm} = v^t \odot a^t$ are the audio/visual channel attention maps. 
% Then, spatial attention captures position-sensitive correspondences between modalities, which is essential for tasks requiring spatial understanding:

\mo{Spatial attention focuses on cross-modal positional importance distributions through modality-specific pooling:}
% \begin{equation}
%     M^t_{vs} = \sigma(\Phi_a(a^t_s)) \in \mathbb{R}^{1\times(H \cdot W)}
% \end{equation}
% \begin{equation}
%     M^t_{as} = \sigma(\Phi_v(v^t_s)) \in \mathbb{R}^{1\times(L\cdot F)}
% \end{equation}
\begin{equation}
     M^t_{vs} = \sigma\left( \Psi_a(a^t) \right) \in \mathbb{R}^{1\times (H \times W)},
\end{equation}
\begin{equation}
     M^t_{as} = \sigma\left( \Psi_v(v^t) \right) \in \mathbb{R}^{1\times (L \times F)},
\end{equation}
\mo{where $\Psi_a(\cdot)$ and $\Psi_v(\cdot)$ implement squeeze-excitation operations to distill spatial dependencies.}
% where $M^t_{vs}$ and $M^t_{as}$ are the A2V and V2A spatial attention maps, respectively.

% Finally, temporal attention addresses the sequential nature of audio-visual data by modeling time-invariant patterns through RNN structures:
\mo{Temporal synchronization is achieved through cascaded RNN layers to model sequential relationships. Given the temporal features $a^t_{tmp}$ and $v^t_{tmp}$ extracted via dense projection:}
\begin{equation}
    M^t_{vt} = \sigma\left( \Gamma_a(RNN(a^t_{tmp})) \right) \in \mathbb{R}^{1\times (H \times W)}, 
\end{equation}
\begin{equation}
    M^t_{at} = \sigma\left( \Gamma_v(RNN(v^t_{tmp})) \right) \in \mathbb{R}^{1\times (H \times W)},
\end{equation}
where $RNN$ is the recurrent neural network to explore the temporal information from video and audio
\mo{, $\Gamma_a$ and $\Gamma_v$ map RNN hidden states to temporal attention scores.}
\mo{The final fused representations integrate the multi-perspective attention through learnable coefficients as formed by:}
% These attention mechanisms are integrated through adaptive weights to form comprehensive cross-modal representations:
% \begin{equation}
%     V^t = (\alpha \cdot M^t_{vc} + \beta \cdot M^t_{vs} + \gamma \cdot M^t_{vt}) \odot V^t
% \end{equation}
% \begin{equation}
%     A^t =(\alpha \cdot M^t_{ac} + \beta \cdot M^t_{as} + \gamma \cdot M^t_{at}) \odot A^t
% \end{equation}
\begin{equation}
\tilde{V}^t = (\alpha \cdot M^t_{vc} \oplus \beta \cdot M^t_{vs} \oplus \gamma \cdot M^t_{vt}) \odot v^t,
% \tilde{A}^t = (\alpha \cdot M^t_{ac} \oplus \beta \cdot M^t_{as} \oplus \gamma \cdot M^t_{at}) \odot a^t,
\end{equation}
\begin{equation}
% \tilde{V}^t = (\alpha \cdot M^t_{vc} \oplus \beta \cdot M^t_{vs} \oplus \gamma \cdot M^t_{vt}) \odot v^t, \quad
\tilde{A}^t = (\alpha \cdot M^t_{ac} \oplus \beta \cdot M^t_{as} \oplus \gamma \cdot M^t_{at}) \odot a^t,
\end{equation}
\mo{where $\oplus$ indicates broadcast addition for dimensional compatibility. This consolidated fusion enables the model to learn generalized cross-modal patterns while preserving task-specific discriminability in shallow layers.}
% where $\alpha$, $\beta$ and $\gamma$ are the hyper-parameters. 
% 通过对视频与音频进行全方位的融合，我们让我们的模型在浅层阶段充分学习多个任务间的音视频通用表征。
% Through the comprehensive fusion of video and audio modalities, our model effectively learns general audio-visual representations across multiple tasks at the shallow phase.

\subsection{Task-specific Modality-shared Dynamic Generating Adapter}
% exp： 为什么不能是模态特定，任务共享
% 1. 任务特定， 模态共享。
% 要优先提及：抗遗忘能力
% 以上是讲共享，这节是讲任务特定，所以一定是抗遗忘能力。
% 大量的可训练共享参数能够为模型带来可观的多任务泛化能力，但同时也会给域差大的任务间带来任务干扰和灾难性遗忘问题。
A large number of trainable shared parameters can bring considerable multi-task generalization capabilities to the model, but at the same time, they may also introduce task interference and catastrophic forgetting issues between tasks with large domain gaps.
% 因此，我们提出了Task-specific Modality-shared Dynamic Generating(TMDG) adapter 来平衡模型的抗遗忘能力与多任务泛化能力
Therefore, we propose the Task-specific Modality-shared Dynamic Generating (TMDG) adapter to balance the model’s anti-forgetting capability with its multi-task generalization ability.
% 特定视听任务中进行audio-visual 特征交互
% 与TMA adapter 不同的是，我们希望TMA能够学习到对于大多数视听任务而言的通用的视听跨模态表征，而我们 设计 TMDG adapter 以引导模型针对于不同的视听任务精炼更深层次的特定跨模态表征。
Unlike the TMA adapter, we aim for TMA to learn a general audio-visual cross-modal representation that is applicable to most audio-visual tasks. In contrast, we designed the TMDG adapter to guide the model in refining deeper task-specific cross-modal representations for different audio-visual tasks.
% 这么做以使得模型取得出色的多任务抗遗忘能力以及特定任务音视频交互能力。
This enables the model to achieve excellent multi-task anti-forgetting capabilities as well as audio-video interaction capabilities for specific tasks.

% Specifically, TMDG adapter aims to select suitable prompts for audio and visual features from a large audio-visual prompt pool. 
\mo{The TMDG operates through a dual-phase process: (1) \textit{prompt pool-based task concept mining} and (2) \textit{dynamic instance-aware prompt generation}.}
% 给定一个预先定义的prompt pool,
\mo{Given a predefined prompt pool $\mathcal{P}= \{p_1, ...,p_L\} \in \mathbb{R}^{L \times d}$, 
the visual feature $\mathcal{V} \in \mathbb{R}^{T\times D}$ and the audio feature $\mathcal{A} \in \mathbb{R}^{T \times D}$ undergo cross-modal fusion through concatenation with prompts, followed by self-attention operations to establish task-specific audio-visual semantic representations. This hierarchical integration process is formally defined as:}
% visual / audio features $V$ /A$ first concatenate with it to achieve the general concept of aaudio visual theme in aa specific task by self-attention, which is formulated by:
% \begin{equation}
%     S_v = \Phi_v(Attn([V, P]))
% \end{equation}
% \begin{equation}
%     S_a = \Phi_v(Attn([A, P]))
% \end{equation}
\begin{equation}
S_v = AvgPool
\left( SelfAttn\left( [\mathcal{V};\mathcal{P}] \right) \right), 
\end{equation}
\begin{equation}
S_a = AvgPool
\left( SelfAttn\left( [\mathcal{A}; \mathcal{P}] \right) \right),
\end{equation}
\mo{where $[\cdot; \cdot]$ denotes concatenation along the token dimension, 
% $S_v$ and $S_a$ are the instance-wise global feature after the temporal average pooling $\Phi_v(\cdot)$ and $\Phi_a(\cdot)$. 
$SelfAttn(\cdot)$ applies multi-head self-attention to model cross-modal dependencies, and $\text{AvgPool}$ consolidates temporal dimensions.}
% 这个特征被缩放并切分为n个头，并以此为引导， prompt pool挑选并生成prompt
Then the instance-wise global feature $S_v$ and $S_a$ are scaled and divided into $n$ parts, guiding the prompt pool to select and generate prompts accordingly, where $n$ is the length of the generated prompts. The formulation can be given as:
\begin{equation}
    G_v = \delta_s(S_v) \cdot P
\end{equation}
\begin{equation}
    G_a = \delta_s(S_a) \cdot P
\end{equation}
where $G_v\in \mathbb{R}^{n \times d}$ and $G_a\in \mathbb{R}^{n \times d}$ denote the generated prompts for visual and audio, respectively. 
$\delta_s$ is a projection layer to scale and divide the instance-wise theme into $n$ parts for different specific prompts.
Finally, the generated prompts are concatenated with the visual and audio features, and then feed into the CLIP and CLAP, respectively.

\subsection{Task-specific Modality-independent Prompts}
% In the deep phase, we develop task-specific modality-independent prompts. 
% These prompts operate in isolation for each task and modality, ensuring the preservation of modality-specific details while excavating profound representations tailored to individual tasks. This strategy reinforces the model’s resilience against the forgetting of information and strengthens its overall performance.

% During audio-visual multi-task learning, the model benefits from mastering general representations across audio and visual modalities, which reveals intrinsic connections and improves its generalization abilities. Nevertheless, an excessive fusion of these modalities risks losing raw information from each, potentially hindering the model’s ability to retain crucial details from one modality in certain tasks. This loss of individual modality information can disrupt the model’s performance on specific tasks that rely on single-modality insights.
% 然而，这些模态的过度融合可能会失去每个模态的原始信息，可能阻碍模型在某些任务中保留某一模态的关键细节。
% 抗遗忘能力和模态特定
% 在深层阶段，为了获取到更加特定于当前任务的音视频特征，且能有效缓解遗忘现象，我们引入了task-specific modality-independent prompts.
In the deep stage, in order to obtain audio-video features more specific to the current task and effectively mitigate the forgetting phenomenon, we introduce Task-specific Modality-Independent~(TMI) prompts.
\mo{While TMA learns shallow cross-modal patterns and TMDG generates task-aware representations, TMI separately preserves modality-critical details through hierarchical prompting.}
% 这个模块以任务特定的prompt为基础，为模型提供任务级的一般表征。
This module is based on task-specific prompts, providing the model with general task-level representations and relieving the issue caused by catastrophic forgetting.
% 此外，类似于AVVP等任务，音视频两者模态间需要独立做出单模态的判断，单一模态的关键细节仍然需要得到保留。
Additionally, for tasks similar to AVVP, it is necessary to make individual modal judgments between the audio and video modalities. The key details of a single modality still need to be preserved.
% 模态独立的必要性
% 模态独立的prompt能够保证模型在单一模态上进一步精炼关键细节特征来引导大模型输出更具任务判别性且更包含模态特点的信息
Modal-independent prompts can ensure that the model refines key detail features on a single modality to guide the large model to output more task-discriminative and modality-characteristic information.
% 深层则能够学习到更复杂的判别性信息和潜在概念以应对不同任务，并且可以更好地处理大规模数据和高复杂度任务。

% Specifically, given the audio feature $A^t$ and visual feature $V^t$ in the \textit{t-th} task, TMI selects the corresponding prompt from different tasks due to the availability of tasks in audio-visual multi-task incremental learning. 
\mo{Given task index $t$ and input features $\mathcal{V}^t \in \mathbb{R}^{T_v \times D}$ and $\mathcal{A}^t \in \mathbb{R}^{T_a \times D}$}, TMI selects the corresponding prompt from different tasks due to the availability of tasks in audio-visual multi-task incremental learning. 
After selecting the visual and audio prompts, we concatenate them with visual and audio features to prompt the large model with a task-wise concept, the formulation is as follows:
\begin{equation}
    PX_v = concat(P_v,V)
\end{equation}
\begin{equation}
    PX_a = concat(P_a,V)
\end{equation}
where $PX_v$ and $PX_a$ denote the new features after being concatenated with the specific prompts $P_v$ and $P_a$. Then prompt-guide features are fed into the large models to achieve a more discriminative clue for the specific task.

\subsection{Optimizing Strategy}
Given $(v_i, a_i,t_i)$ as a piece of visual-audio-text pair indexed by $i$ in the \textit{d-th} task, the visual embedding $F^v_i$, the audio embedding $F^a_i$ and the corresponding text embedding $T^v_i$ and $T^a_i$ is encoded by:
\begin{equation}
    F^v_i=MLP^v(\phi_v(v_i)) \qquad  F^a_i = MLP^a(\phi_a(a_i))
\end{equation}
\begin{equation}
    T^v_i=MLP^v_t(\phi^v_t(t_i)) \qquad  F^a_i = MLP^a_t(\phi^a_t(t_i))
\end{equation}
where $MLP$s are the 2-layer multilayer perception with ReLU for visual/audio/text respectively to project them into the same dimension.
Then the model employs the contrastive constraint between the paired visual and text embeddings, followed by CLIP. Similar to CLIP, the paired audio and text embeddings conduct the contrastive constraints followed by CLAP, the formulations can be written as:
\begin{equation}
\begin{aligned}
    \mathcal{L}_v = & \frac{1}{2N} \sum^N_{i=1}(log\frac{exp(F_i^v \cdot T^v_i /\tau_v)}{\sum^N_{j=1}{exp(F_i^v \cdot T^v_j/ \tau_v)}} \\
    &+ log\frac{exp(T^v_i \cdot F_i^v /\tau_v)}{\sum^N_{j=1}{exp(T^v_i \cdot F_j^v/ \tau_v)}})
\end{aligned}   
\end{equation}
\begin{equation}
\begin{aligned}
    \mathcal{L}_a = & \frac{1}{2N} \sum^N_{i=1}(log\frac{exp(F_i^a \cdot T^a_i /\tau_a)}{\sum^N_{j=1}{exp(F_i^a \cdot T^a_j/ \tau_a)}} \\
    &+ log\frac{exp(T^a_i \cdot F_i^a /\tau_a)}{\sum^N_{j=1}{exp(T^a_i \cdot F_j^a/ \tau_av)}})
\end{aligned}   
\end{equation}
where $\tau_v$ and $\tau_a$ are learnable temperature parameters to adjust the loss. $N$ represents the batch size, $exp(\cdot)$ denote the exponential function.

%% file: sec/4_experiments.tex
\section{Experiments}
\mo{In this work, we simulate audio-visual multi-task incremental learning by treating multiple audio-visual tasks as a continuous data stream. 
% 在这个过程中，我们采用了多个不同的视听理解类任务，包括 Audio-Visual Event localization， Audio-Visual Video Parsing 和Audio-Visual Question Answering
We employ multiple different audio-visual understanding tasks, including audio-visual event localization~(AVE), audio-visual video parsing~(AVVP), audio-visual question answering~(AVQA) and audio-visual segmentation~(AVS). Detailed experimental settings can be found in the \textbf{supplementary material}.}

\begin{table*}[]
  \vspace{-1.5em}
  \centering
    \setlength{\abovecaptionskip}{0.01cm}
    \setlength{\belowcaptionskip}{0.01cm}
  \caption{Comparison on anti-forgetting ability with the SOTA methods. $\uparrow$ indicates higher is better, $\downarrow$ indicates lower is better. The best results are highlighted in bold.}
  \resizebox{\textwidth}{!}{
    \begin{tabular}{l|ccc|ccc|ccc|ccc}
    \toprule[1.5pt]
    \multicolumn{1}{c|}{\multirow{2}[2]{*}{\textbf{Methods}}} & \multicolumn{3}{c|}{\textbf{AVE}} & \multicolumn{3}{c|}{\textbf{AVVP}} & \multicolumn{3}{c|}{\textbf{AVQA}} & \multirow{2}[2]{*}{\textbf{$\overline{A}_{mean}$ ↑}} & \multirow{2}[2]{*}{\textbf{$\overline{F}_{mean}$ ↓}} & \multirow{2}[2]{*}{\textbf{$\overline{A}_{final}$ ↑}} \\
          & \textbf{$A_{mean}$ ↑} & \textbf{$A_{final}$ ↑} & \textbf{$F_{mean}$ ↓} & \textbf{$A_{mean}$ ↑} & \textbf{$A_{final}$ ↑} & \textbf{$F_{mean}$ ↓} & \textbf{$A_{mean}$ ↑} & \textbf{$A_{final}$ ↑} & \textbf{$F_{mean}$ ↓} &       &       &  \\
    \midrule
    \rowcolor{gray!10} Fine-tune & 29.61 & 12.74  & 22.37  & \textbf{48.52} & 38.30  & 7.17 & 53.66 & 53.49  & 0.35  & 43.93 & 9.96  & 34.84 \\
    EWC   & 8.52  & 2.38  & 7.71  & 22.55 & 1.08  & 30.97 & 56.73 & 56.70 & 0.00  & 29.26 & 12.89 & 20.05 \\
    L2P   & 52.05 & 34.25 & 17.78 & 41.43 & 33.23 & 7.75  & 59.77 & 59.80 & -0.01 & 51.08 & 8.51  & 42.42 \\
    \rowcolor{gray!10}S-prompt & 57.58 & 52.75 & \textbf{4.78}  & 40.36 & 30.13 & 9.15  & 59.17 & 59.21 & -0.04 & 52.37 & 4.63  & 47.36 \\
    Dualprompt & \textbf{63.00} & \textbf{58.10} & 5.03  & 37.31 & 28.30 & 8.81  & 63.39 & 63.47 & -0.08 & 54.57 & 4.59  & 49.95 \\
    \rowcolor{gray!10} PC    & 62.02 & 54.44 & 7.62  & 39.07 & 30.98 & 7.03  & 69.55 & 69.46 & 0.04  & 56.88 & 4.90  & 51.63 \\
    DCNet & 30.43  & 12.75 & 23.29   & 48.75  & 39.17  & 6.52   & 53.34  & 53.86  & 0.16  & 44.17 & 9.99  & 35.26 \\
    \midrule
    \rowcolor{blue!15} \textbf{Ours} & 62.03 & 55.00 & 6.72  & 45.25 & \textbf{40.67} & \textbf{2.22}  & \textbf{69.29} & \textbf{68.56} & 1.01  & \textbf{58.85} & \textbf{3.32} & \textbf{54.74} \\
    \bottomrule[1.5pt]
    \end{tabular}}
  \label{tab_com_1}%
\end{table*}%

\begin{table*}[]
 \centering
    \setlength{\abovecaptionskip}{0.01cm}
    \setlength{\belowcaptionskip}{0.01cm}
 \caption{Comparison on multi-task transfer ability with the SOTA methods. $A_{single}$ and $A_{multi}$ represent the performance of single-task and multi-task training, respectively. The best results are highlighted in bold and Diff indicates the normalized improvement with quadratic baseline penalty. (Details can be seen in supplementary.)}
 \scalebox{0.8}{
 \setlength{\tabcolsep}{3.2mm}% 调整表宽度(如果需要)
 % \resizebox{440pt}{!}{
   \begin{tabular}{l|c|cc|cc|cc|cc|cc}
   \toprule[1.5pt]
   \multicolumn{1}{c|}{\multirow{2}[2]{*}{\textbf{Methods}}} & \multirow{2}[2]{*}{\textbf{Venue}} & \multicolumn{2}{c|}{\textbf{AVE}} & \multicolumn{2}{c|}{\textbf{AVVP}} & \multicolumn{2}{c|}{\textbf{AVQA}} & \multirow{2}[2]{*}{$\overline{A}_{single}$} & \multirow{2}[2]{*}{$\overline{A}_{multi}$} & \multirow{2}[2]{*}{\textbf{Diff}} \\
         &       & $A_{single}$ ↑ & $A_{multi}$ ↑ & $A_{single}$ ↑ & $A_{multi}$ ↑ & $A_{single}$ ↑ & $A_{multi}$ ↑ &       &   &  \\
   \midrule
   \rowcolor{gray!10} Fine-tune & - & 57.47  & 18.22  & 52.64  & 59.00  & 54.19  & 54.13  & 54.77 & 43.78 & {\color{red!60!black}-58.16\%} \\
   EWC   & PNAS'17 & 17.79 & 18.00 & 63.01 & 62.67 & 56.70 & 54.64 & 45.83 & 45.10 & {\color{red!60!black}-2.87\%} \\
   L2P   & CVPR'22 & 69.80 & 67.55 & 48.72 & 42.84 & 59.79 & 60.04 & 59.44 & 56.81 & {\color{red!60!black} -16.46\%} \\
   \rowcolor{gray!10} S-prompt & NeurIPS'22 & 62.31 & 54.39 & 48.42 & 42.84 & 59.13 & 54.57 & 56.62 & 50.60 & {\color{red!60!black}-34.01\%} \\
   Dualprompt & ECCV'22 & 68.15 & 67.48 & 45.92 & 46.19 & 63.32 & 60.04 & 59.13 & 57.90 & {\color{red!60!black}-7.59\%} \\
   \rowcolor{gray!10} PC    & ACM MM'24 & 69.67 & 68.04 & 45.04 & 44.61 & 69.54 & 69.85 & 61.41 & 60.83 & {\color{red!60!black} -3.94\% } \\
   DCNet & ARXIV'25 & 59.33 & 19.06  & 52.20  & 57.60  & 54.18  & 53.97  & 55.24 & 43.54 & {\color{red!60!black}-62.98\%} \\
   \midrule
   \rowcolor{blue!15} \textbf{Ours} & - & \textbf{70.45} & \textbf{72.52} & 47.80 & \textbf{48.60} & 68.84 & 69.30 & \textbf{62.36} & \textbf{63.47} & {\color{green!60!black}\textbf{+7.79\%}} \\
   \bottomrule[1.5pt]
   \end{tabular}}
 \label{tab_com_2}%s
 \vspace{-1.5em}
\end{table*}%

\subsection{Comparison Results}
% 描述这个小结的用途，做什么
In this section, we conduct the comparison experiments between the proposed method PHP and the SOTA methods. The following methods can be divided into three groups: 
% 这个方法应对新到来的任务，采用在前序任务的基础上进一步对模型进行微调
1) Fine-tune-based method. This method addresses the newly arrived tasks by further fine-tuning the model based on the previous tasks.
2) Regularization-based method including EWC~\cite{kirkpatrick2017overcoming}.
3) Prompt-based methods including L2P~\cite{l2p}, S-prompt~\cite{sprompt}, Dualprompt~\cite{dualprompt}, PC~\cite{pc}, DCNet~\cite{dcn}.
For fairness, the same pre-trained models~(\textit{i.e.} CLIP and CLAP) are used for all compared models as well as ours.

\begin{table}[tp]
  \centering
    \setlength{\abovecaptionskip}{0.01cm}
    \setlength{\belowcaptionskip}{0.01cm}
  \caption{Ablation study on anti-forgetting ability. }
    \resizebox{\linewidth}{!}{
    \begin{tabular}{l|ccc|ccc}
    \toprule[1.5pt]
    \multirow{2}[2]{*}{\textbf{Row}} & \textbf{TMA} & \textbf{TMDG} & \textbf{TMI} & \multirow{2}[2]{*}{$\overline{A}_{mean}$ ↑} & \multirow{2}[2]{*}{$\overline{A}_{final}$ ↑} & \multirow{2}[2]{*}{$\overline{F}_{mean}$ ↓} \\
          & \textbf{adapter} & \textbf{adapter} & \textbf{prompts} &       &       &  \\
    \midrule
    \rowcolor{gray!10} 1     &   &   &   & 58.20 & 54.22 & 4.20 \\
    2     & \Large \checkmark  &   &   & 58.23 & 55.07 & 3.34 \\
    \rowcolor{gray!10} 3     &   & \Large \checkmark  &   & 59.20 & 55.41 & 3.39 \\
    4     &   &   & \Large \checkmark  & 59.70 & 55.84 & 3.78 \\
    \rowcolor{gray!10} 5     & \Large \checkmark  & \Large \checkmark  &   & 59.54 & 56.01 & 3.21 \\
    6     & \Large \checkmark  &   & \Large \checkmark  & 58.87 & 55.20 & 3.76 \\
    \rowcolor{gray!10} 7     &   & \Large \textbf{\checkmark} & \Large \checkmark  & 59.28 & 45.51 & 3.90 \\
    8    & \Large \checkmark  & \Large \checkmark  & \Large \checkmark  & 58.85 & 54.74  & 3.32  \\
    \bottomrule[1.5pt]
    \end{tabular}}%
  \label{abl_1}%
  \vspace{-1.5em}
\end{table}%

% 比较 
\noindent
\textbf{Comparison on anti-forgetting ability with the SOTA methods.}
% 在增量任务中，抗遗忘问题是一个经典且亟需解决的问题
Catastrophic forgetting is a key problem in incremental learning, and solving it is essential for models to remember old information while learning new tasks. 
% 在这个实验中，我们通过和sota方法比较来证明我们的模型的抗遗忘能力
In this experiment, we demonstrate the anti-forgetting capability of our model by comparing it with the SOTA methods.
Table~\ref{tab_com_1} shows the comparison on three tasks, including AVE, AVVP and AVQA. 
The experiment is conducted on different incremental orders~(the detailed results can be seen in the supplementary), and then we compute the mean accuracy and forget of different tasks. $\overline{A}_{mean}$, $\overline{F}_{mean}$ and $\overline{A}_{final}$ are computed as mean accuracy, mean forgetting scores and final accuracy by all tasks, respectively.

% Particularly, it outperforms the proposal-based method MS-2D-TAN by 2.34\% and 6.78\% relative improvement in terms of $R@1, IoU=0.5$ and $R@1, IoU=0.7$, respectively.
As shown in Table~\ref{tab_com_1}, our approach outperforms the other methods in terms of all metrics. Specifically, we notice that the mean accuracy is higher than that of other methods, which shows that our method performs better on audio-visual tasks. 
The mean forgetting score is 1.58\% lower than the SOTA prompt-based incremental method, which demonstrates our model has a more significant ability to mitigate the forgetting issue resulting from the incremental process. The outstanding mean of the final accuracy also illustrates that our model can effectively maintain the initial performance after training with two different tasks. 

% 总结。

\begin{table}[!b]
  \centering
  \vspace{-1.5em}
     \setlength{\abovecaptionskip}{0.01cm}
     \setlength{\belowcaptionskip}{0.01cm}
  \caption{Ablation study on transfer ability.}
    \resizebox{\linewidth}{!}{
    \begin{tabular}{l|ccc|cc|c}
    \toprule[1.5pt]
    \multirow{2}[2]{*}{\textbf{Row}} & \textbf{TMA} & \textbf{TMDG} & \textbf{TMI} & \multirow{2}[2]{*}{$\overline{A}_{single}$} & \multirow{2}[2]{*}{$\overline{A}_{multi}$} & \multirow{2}[2]{*}{\textbf{Diff}} \\
          & \textbf{adapter} & \textbf{adapter} & \textbf{prompts} &       &   &  \\
    \midrule
    \rowcolor{gray!10} 1     &   &   &   & 62.62 & 61.76 & {\color{red!60!black}-6.10\%} \\
    2     & \Large \textbf{\checkmark}  &   &   & 61.75 & 62.93 & {\color{green!60!black}+8.05\%} \\
    \rowcolor{gray!10} 3     &   & \Large \textbf{\checkmark}  &  & 62.20 & 62.14 & {\color{red!60!black}-0.37\%} \\
    4     &   &   & \Large \textbf{\checkmark}  & 63.41 & 63.34 & {\color{red!60!black}-0.48\%} \\
    \rowcolor{gray!10} 5     & \Large \textbf{\checkmark}  & \Large \textbf{\checkmark}  &   & 62.44 & 63.12 & {\color{green!60!black}+4.77\%} \\
     6     & \Large \textbf{\checkmark}  &   & \Large \textbf{\checkmark}  & 62.73 & 64.26 & {\color{green!60!black}+7.99\%} \\
    \rowcolor{gray!10}7     &   & \Large \textbf{\checkmark}  & \Large \textbf{\checkmark}  & 63.31 & 63.21 & {\color{red!60!black}-0.71\%} \\
    8    & \Large \textbf{\checkmark}  & \Large \textbf{\checkmark}  & \Large \textbf{\checkmark}  & 62.36 & 63.47 & {\color{green!60!black}+7.77\%} \\
    \bottomrule[1.5pt]
    \end{tabular}}%
  \label{abl_2}%
  \vspace{-1.5em}
 \end{table}%

\noindent
\textbf{Comparison on multi-task transfer ability with the SOTA methods.}
% 泛化
% The experiment mainly proves the transfer ability of our model by comparing it with the SOTA methods.
% As shown in Table~\ref{tab_com_2}, $\overline{A}_{single}$ represents the average of accuracy with individual task training of three tasks, while $\overline{A}_{multi}$ denotes the average of accuracy after training two audio-visual tasks. 
% The performance of $\overline{A}_{multi}$ is higher than that of $\overline{A}_{single}$, which indicates that the model has good transferability. It can leverage the previous tasks to assist the current task in learning and improving.
% Specifically, our model surpasses the fine-tuning method by a large margin. Although the fine-tuning method has outstanding performance in knowledge transfer, its massive learnable parameters also cause serious task interference problems. Our method achieves outstanding results by balancing task-specific and task-shared components. Compared with the SOTA prompt-based incremental method, PC, our model outperforms 0.95\% and 2.64\% in terms of $\overline{A}_{multi}$ and $\overline{A}_{single}$, respectively. The observation reveals that while the PC demonstrates an outstanding capacity for retention, it falls short in terms of transferability. 
% % 我们猜测，由于任务特定参数的过多导致了模型对任务间共通表征不敏感，从而抑制了模型的迁移能力。
% We speculate that the excessive number of task-specific parameters leads to the model being insensitive to the shared representations between tasks, thereby inhibiting its transferability.
% % 需要总结吗？ mmm
The experiment primarily demonstrates the transfer ability of our model by comparing it with the state-of-the-art methods. As shown in Table~\ref{tab_com_2}, our method achieves the highest single-task performance (62.36\%) and multi-task performance (63.47\%) across all compared approaches. Most importantly, while all baseline methods exhibit negative transfer (ranging from -2.87\% to -62.98\%), our method is the only one showing positive transfer ability with a significant improvement of +7.79\%. This indicates that our progressive prompting approach effectively leverages knowledge from previously learned tasks to benefit new tasks. Specifically, on the AVE task, our method surpasses others with both the highest single-task accuracy (70.45\%) and a notable improvement after multi-task learning (72.52\%). For AVVP and AVQA tasks, our approach consistently maintains strong performance, avoiding the performance degradation that affects competing methods. These results validate the effectiveness of our hierarchical design in balancing task-specific learning and cross-task knowledge transfer.

\begin{table}[!b]
  \centering
  \setlength{\abovecaptionskip}{0.01cm}
  \setlength{\belowcaptionskip}{0.01cm}
  \caption{Ablation study on different orders of TMA adapter, TMDG adapter, and TMI prompts.}
  \resizebox{\linewidth}{!}{
    \begin{tabular}{c|ccc|ccc}
    \toprule[1.5pt]
    Row & \begin{tabular}[c]{@{}c@{}}TMA\\adapter\end{tabular} & \begin{tabular}[c]{@{}c@{}}TMDG\\adapter\end{tabular} & \begin{tabular}[c]{@{}c@{}}TMI\\prompts\end{tabular} & $A_{mean}$ $\uparrow$ & $A_{final}$ $\uparrow$ & $F_{mean}$ $\downarrow$ \\
    \midrule
    \rowcolor{gray!10} 1 & D & M & S & 47.29 & 41.14 & 7.29 \\
    2 & M & D & S & 48.85 & 43.32 & 5.42 \\
    \rowcolor{gray!10} 3 & D & S & M & 48.32 & 44.17 & 5.89 \\
    4 & M & S & D & 49.12 & 42.73 & 7.41 \\
    \rowcolor{gray!10} 5 & S & D & M & 47.82 & 43.08 & 5.70 \\
    6 & S & M & D & \textbf{58.85} & \textbf{54.74}  & \textbf{3.32} \\
    \bottomrule[1.5pt]
    \end{tabular}}%
  \label{tab:ablation_order_forget}%
 \end{table}

 \begin{table}[!b]
  \centering
    \setlength{\abovecaptionskip}{0.01cm}
    \setlength{\belowcaptionskip}{0.01cm}
  \caption{Ablation study on transfer ability with different orders of components.}
  \resizebox{\linewidth}{!}{
    \begin{tabular}{c|ccc|ccc}
    \toprule[1.5pt]
    Row & \begin{tabular}[c]{@{}c@{}}TMA\\adapter\end{tabular} & \begin{tabular}[c]{@{}c@{}}TMDG\\adapter\end{tabular} & \begin{tabular}[c]{@{}c@{}}TMI\\prompts\end{tabular} & $A_{single}$ $\uparrow$ & $A_{multi}$ $\uparrow$ & \textbf{Diff} \\
    \midrule
    \rowcolor{gray!10} 1 & D & M & S & 55.72 & 55.43 &  {\color{red!60!black}-1.55\%}\\
    2 & M & D & S & 54.17 & 55.38 & {\color{green!60!black}+6.30\%} \\
    \rowcolor{gray!10} 3 & D & S & M & 55.95 & 57.08 & {\color{green!60!black}+6.20\%} \\
    4 & M & S & D & 57.55 & 56.60 & {\color{red!60!black}-5.57\%} \\
    \rowcolor{gray!10} 5 & S & D & M & 54.47 & 54.98 & {\color{green!60!black}+2.69\%} \\
    6 & S & M & D & \textbf{62.36} & \textbf{63.47} & {\color{green!60!black}+7.77\%} \\
    \bottomrule[1.5pt]
    \end{tabular}}%
  \label{tab:ablation_order_transfer}%
 \end{table}

\subsection{Ablation Studies}
In this section, we conduct the ablation experiments to prove the effectiveness of each module or task order.
% 整体消融实验还是有点问题，目前模型中的一些问题已经排除，涉及到的问题得重跑。

\noindent
\textbf{Ablation study on anti-forgetting ability.}
Table~\ref{abl_1} presents the ablation results on anti-forgetting performance. The baseline (Row 1) shows moderate performance with mean forgetting of 4.20\%. 
Each component makes unique contributions: TMI prompts achieve the highest average accuracy 
($\overline{A}_{mean}=59.70\%$) and final accuracy while the TMA+TMDG combination (Row 5) yields the lowest forgetting rate ($\overline{F}_{mean}=3.21\%$) and highest final accuracy ($\overline{A}_{final}= 56.01\%$). 
This demonstrates how different components contribute to maintaining performance and mitigating catastrophic forgetting in audio-visual incremental learning.

% \noindent
% \textbf{Ablation study on transfer ability.}
% % 缺少 对于实验的描述
% Table~\ref{abl_2} analyzes the transfer ability through ablation studies. The baseline shows negative transfer with a -4.60\% decline. While TMI prompts alone achieves the highest single-task performance at 63.41\%, it shows a slight negative transfer at -0.13\%. Notably, combining TMA adapter with TMI prompts achieves optimal multi-task performance of 64.09\% with positive transfer at +6.00\%, demonstrating the necessity of balanced task-shared and task-specific features for effective knowledge transfer.

\noindent
\textbf{Ablation study on transfer ability.}
Table \ref{abl_2} analyzes each component's contribution to knowledge transfer. Without any components (Row 1), the model shows significant negative transfer (-6.10\%). The TMA adapter alone (Row 2) achieves strong positive transfer (+8.05\%), highlighting the importance of universal cross-modal representations. The TMA+TMI configuration (Row 6) and full model (Row 8) deliver similarly strong transfer capabilities (+7.99\% and +7.77\%). Interestingly, TMDG+TMI (Row 7) shows negative transfer (-0.71\%), indicating that without TMA's foundational representations, higher-level components cannot effectively share knowledge across tasks.

\noindent
\mo{\textbf{Ablation study on different component orders. }Tables~\ref{tab:ablation_order_forget} and \ref{tab:ablation_order_transfer} present a comprehensive analysis of different component arrangements, where S indicates placement in shallow layers, M in middle layers, and D in deep layers. As shown in Table 5, the "S-M-D" configuration (our proposed progressive arrangement) significantly outperforms alternative orderings with the highest mean accuracy (58.85\%) and lowest forgetting rate (3.32\%). Table 6 further demonstrates that this progressive ordering achieves optimal transfer capability with the highest single-task (62.36\%) and multi-task (63.47\%) performances. In contrast, configurations with "D" or "M" components in shallow layers exhibit higher forgetting rates and negative transfer effects. These results strongly validate our progressive design principle, where universal representations are learned at shallow layers, task-specific cross-modal features at middle layers, and fine-grained task-modality details at deeper layers.}

% 调参
\begin{figure}[tp]
    \centering
    % \vspace{-1.5em}
    % \setlength{\abovecaptionskip}{0.01cm}
    % \setlength{\belowcaptionskip}{0.01cm}
    \includegraphics[scale=0.36]{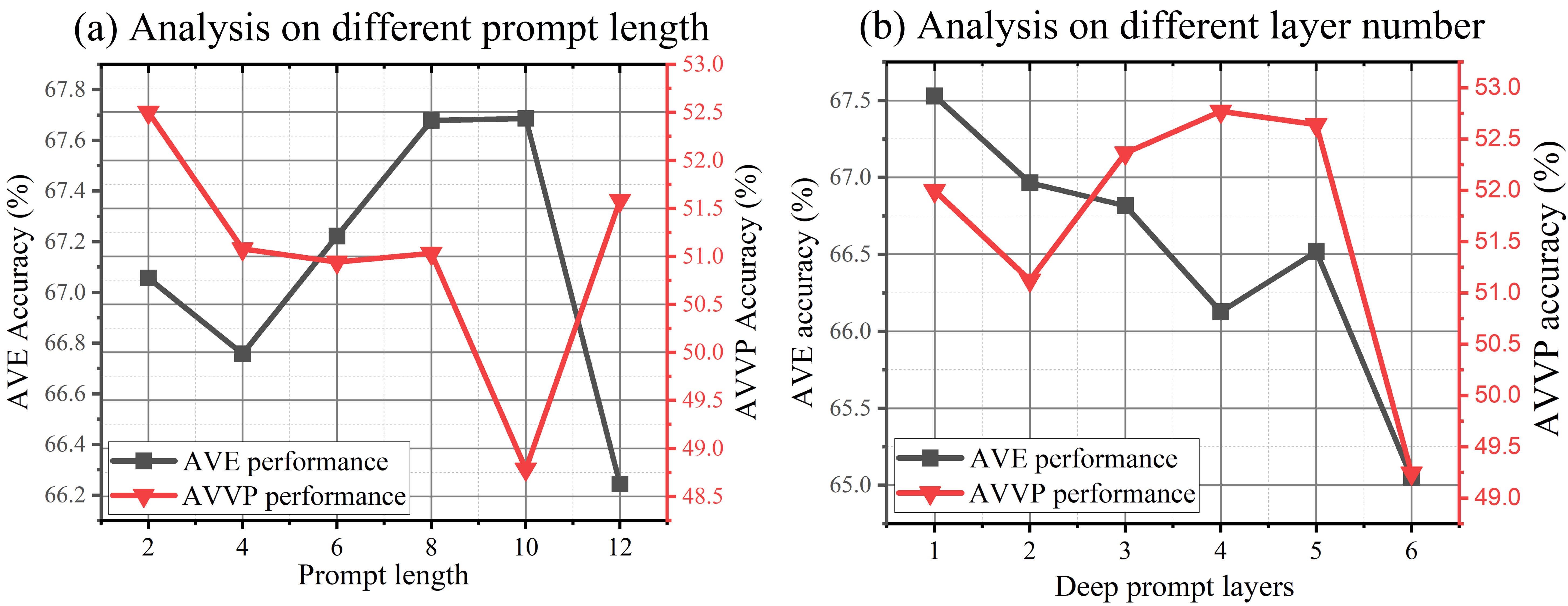} % 0.63  0.67
    \caption{
    Ablation studies on key design parameters. (a) Impact of prompt length on model performance. (b) Effect of task-specific layer depth on AVE and AVVP performance.
    }
    \label{fig_para}
    % \vspace{-10pt}
    % \vspace{-1.5em}
\end{figure}

\begin{figure}[!t]
    \centering
    % \vspace{-1.5em}
    % \setlength{\abovecaptionskip}{0.01cm}
    % \setlength{\belowcaptionskip}{0.01cm}
    \includegraphics[scale=0.20]{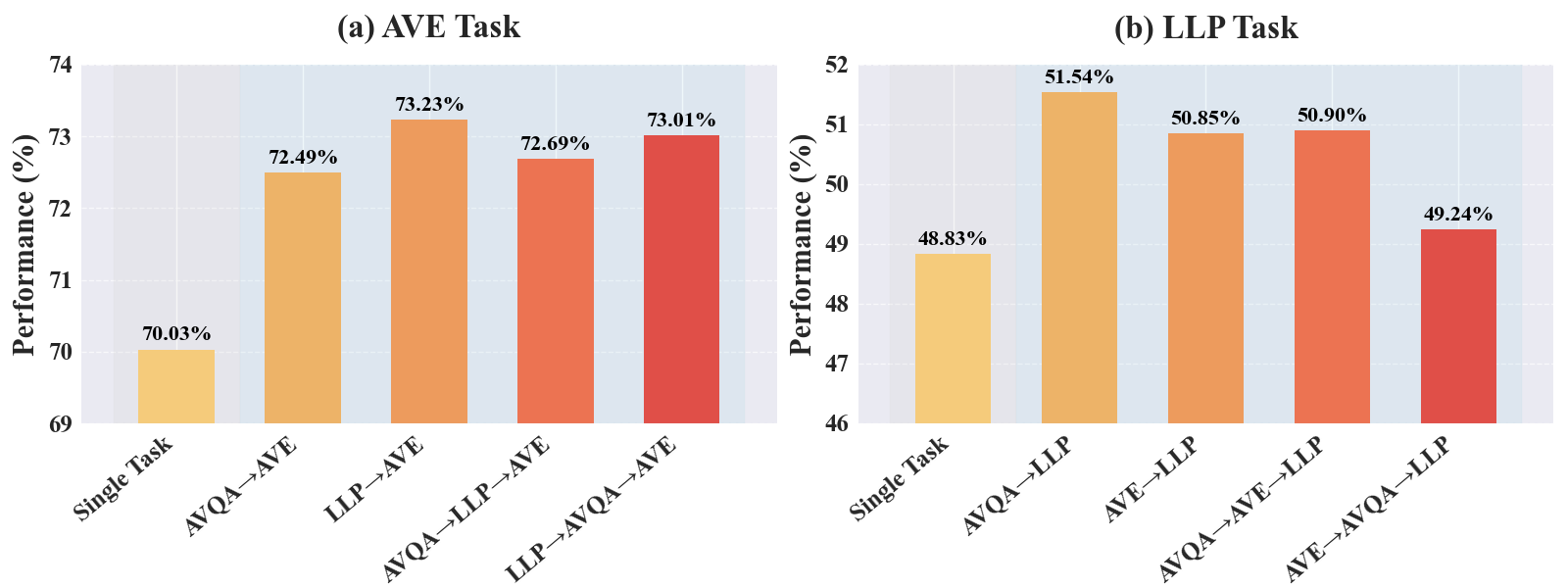} % 0.63  0.67
    \caption{
    Performance analysis of multi-task training sequences. Comparison of different training orders on (a) AVE task and (b) LLP task performance. 
    }
    \label{fig_multi_single}
    % \vspace{-10pt}
    \vspace{-1.5em}
\end{figure}

\begin{figure}[bp]
    \centering
    \vspace{-1em}
    \setlength{\abovecaptionskip}{0.01cm}
    \setlength{\belowcaptionskip}{0.01cm}
    \includegraphics[scale=0.41]{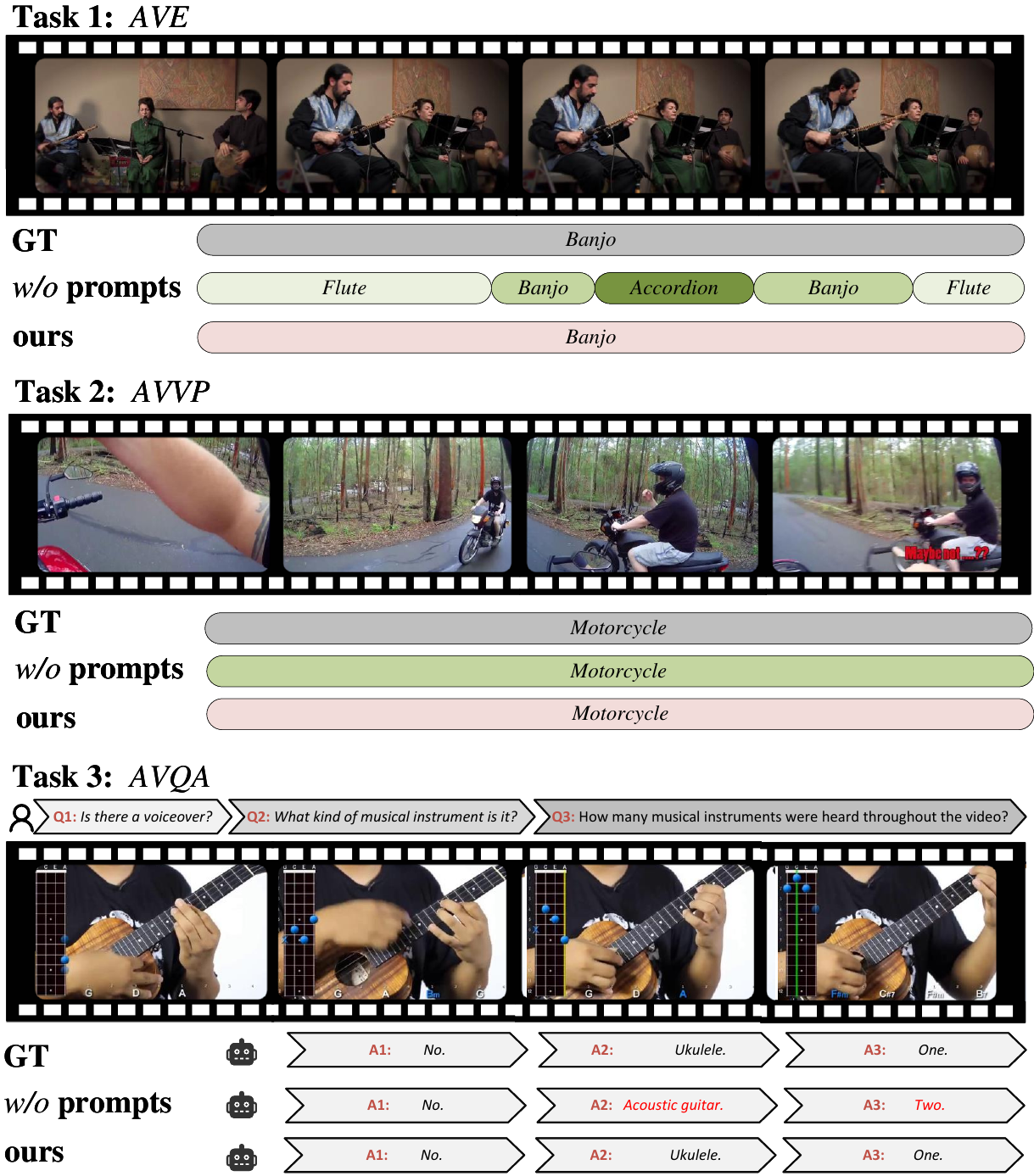} % 0.63  0.67
    \caption{
    Qualitative Results. We compare our method with baseline (without prompts) on AVE, AVVP, and AVQA. Our method demonstrates superior performance in maintaining accurate cross-modal understanding and temporal localization.
    }
    \label{fig_vis}
    % \vspace{-10pt}
    % \vspace{-1.5em}
\end{figure}

% \begin{table}[tp]
%   \centering
%   \caption{Comparison of different module orders on model performance metrics.}
%   \resizebox{\linewidth}{!}{
%     \begin{tabular}{c|ccc|cc}
%     \toprule[1.5pt]
%      \textbf{Order} & $A_{mean}$ $\uparrow$ & $A_{final}$ $\uparrow$ & $F_{mean}$ $\downarrow$ & $A_{single}$ $\uparrow$ & $A_{multi}$ $\uparrow$ \\
%     \midrule
%     \rowcolor{gray!10} dms & 47.29 & 41.14 & 7.29 & 55.72 & 55.43 \\
%     dsm & 48.85 & 43.32 & 5.42 & 54.17 & 57.17 \\
%     \rowcolor{gray!10} mds & 48.32 & 44.17 & 5.89 & 55.95 & 57.08 \\
%     msd & 49.12 & 42.73 & 7.41 & 57.55 & 56.60 \\
%     \rowcolor{gray!10} sdm & 47.82 & 43.08 & 5.70 & 54.47 & 54.98 \\
%     \midrule
%     \rowcolor{blue!15} smd & \textbf{57.92} & \textbf{53.43} & \textbf{4.61} & \textbf{62.66} & \textbf{63.18} \\
%     \bottomrule[1.5pt]
%     \end{tabular}}%
%   \label{tab:performance_metrics}%
%  \end{table}

\noindent
\textbf{Ablation study on deep prompts.}
\textbf{Ablation study on deep prompts.}
We investigate how prompt length and number of injected layers affect model performance using the \textit{AVE→AVQA→AVVP} sequence. We track AVE performance to evaluate forgetting resistance and AVVP performance to assess transfer capability.
As shown in Fig.~\ref{fig_para}(a), increasing prompt length improves resistance to forgetting but reduces knowledge transfer efficiency, revealing a clear trade-off between memory retention and cross-task generalization.
Fig.~\ref{fig_para}(b) demonstrates that increasing task-specific layers gradually degrades AVE performance, while AVVP performance first improves then sharply declines beyond the optimal configuration. This indicates that moderate task-specific parameterization helps capture relevant features, but excessive parameters harm knowledge transferability across tasks.

\noindent
\textbf{Ablation study between single-task and multi-task.}
Fig.~\ref{fig_multi_single} presents the performance analysis of different training sequences on AVE and LLP tasks. For AVE task (a), the model achieves 70.03\% accuracy in single-task training, while all incremental sequences show improved performance, with LLP→AVE achieving the highest at 73.23\%. For LLP task (b), AVQA→LLP sequence performs best at 51.54\%, significantly outperforming the single-task baseline of 48.83\%. The results demonstrate that appropriate task ordering in incremental learning can enhance model performance beyond single-task training.

\noindent
\textbf{Incremental learning on Four-Task Sequence.}
We further investigate our method on longer task sequences with four consecutive audio-visual tasks. As shown in Table~\ref{tab:four_task_sequences}, our approach shows remarkable stability across different task orderings. For the AVS→AVQA→AVVP→AVE sequence, the model maintains consistent performance on the initial task (55.51\% final accuracy) while successfully learning subsequent tasks. Similarly, in the AVE→AVVP→AVS→AVQA order, our method preserves strong performance on earlier tasks while effectively transferring knowledge to new domains. These results validate that our progressive prompting strategy scales effectively to longer task sequences while maintaining a favorable balance between knowledge retention and transfer capability.

\subsection{Quantitative Results}
To provide an intuitive understanding of our method's effectiveness, we visualize results across three audio-visual tasks in Fig~\ref{fig_vis}. For the AVE task, our method accurately identifies "Banjo" segments, while the baseline confuses it with "Flute" and "Accordion", demonstrating our model's superior cross-modal feature alignment. In AVVP, both approaches correctly detect motorcycle sounds, but our method achieves more precise temporal localization, validating the effectiveness of our task-specific adapter design. For AVQA, our approach correctly answers all questions about the musical instrument (ukulele), while the baseline misidentifies it as an acoustic guitar and incorrectly counts the number of instruments. These qualitative results not only demonstrate our model's superior ability to maintain accurate cross-modal understanding but also highlight its effectiveness in preserving task-specific features through incremental learning. The consistent performance across diverse tasks further validates the robustness of our work.

% % Table generated by Excel2LaTeX from sheet 'comparison'
% \begin{table}[tp]
%  \centering
% % \setlength{\abovecaptionskip}{0.01cm}
% % \setlength{\belowcaptionskip}{0.01cm}
%  \caption{Detailed performance under different task orders. Each column within an order shows the performance of tasks as they are incrementally learned. The first task's performance is tracked across all stages to evaluate forgetting resistance, while the final task's performance demonstrates transfer capability.}
%  \resizebox{\linewidth}{!}{
%    \begin{tabular}{l|ccc|ccc|ccc}
%    \toprule[1.5pt]
%    \rowcolor{blue!15} Stage & \multicolumn{3}{c|}{AVE→AVVP→AVQA} & \multicolumn{3}{c|}{AVE→AVQA→AVVP} & \multicolumn{3}{c}{AVVP→AVE→AVQA} \\
%    \midrule
%    \rowcolor{gray!10} 1     & 70.62 &       &       & 70.03 &       &       & 50.44 &       &  \\
%    2     & 60.52 & 49.38 &       & 70.22 & 70.46 &       & 42.22 & 71.39 &  \\
%    \rowcolor{gray!10} 3     & 58.23 & 53.05 & 70.26 & 56.39 & 70.02 & 49.24 & 35.98 & 69.58 & 70.11 \\
%    \midrule
%    \rowcolor{blue!15} Stage & \multicolumn{3}{c|}{AVVP→AVQA→AVE} & \multicolumn{3}{c|}{AVQA→AVE→AVVP} & \multicolumn{3}{c}{AVQA→AVVP→AVE} \\
%    \midrule
%    \rowcolor{gray!10} 1     & 50.07 &       &       & 69.72 &       &       & 69.93 &       &  \\
%    2     & 49.47 & 69.74 &       & 69.62 & 72.69 &       & 69.64 & 50.04 &  \\
%    \rowcolor{gray!10} 3     & 28.36 & 69.62 & 73.01 & 69.71 & 62.16 & 50.90 & 68.34 & 39.06 & 71.19 \\
%    \bottomrule[1.5pt]
%    \end{tabular}}%
%  \label{tab:task_order_comparison}%
%  % \vspace{-1.5em
% \end{table}%

\begin{table}[tp]
  \centering
  \setlength{\abovecaptionskip}{0.01cm}
  \setlength{\belowcaptionskip}{0.01cm}
  \caption{Performance across different four-task incremental learning sequences.}
  \resizebox{\linewidth}{!}{
    \begin{tabular}{l|cccc|cccc}
    \toprule[1.5pt]
    \multicolumn{9}{c}{\textbf{Four-Task Incremental Learning}} \\
    \midrule
    \rowcolor{blue!15} Stage & \multicolumn{4}{c|}{AVS→AVQA→AVVP→AVE} & \multicolumn{4}{c}{AVE→AVVP→AVS→AVQA} \\
    \midrule
    \rowcolor{gray!10} 1     & 58.62 &       &       &       & 70.77 &       &       &       \\
    2     & 54.25 & 69.36 &       &       & 60.17 & 48.00 &       &       \\
    \rowcolor{gray!10} 3     & 54.28 & 66.31 & 48.60 &       & 60.60 & 44.42 & 58.82 &       \\
    4     & 55.51 & 66.61 & 31.48 & 69.88 & 59.58 & 50.90 & 53.19 & 69.80 \\
    \bottomrule[1.5pt]
    \end{tabular}}%
  \label{tab:four_task_sequences}%
  \vspace{-1.5em}
 \end{table}%

%% file: sec/5_conclusion.tex
\section{Conclusion}
In this paper, we present progressive homeostatic and plastic prompts for audio-visual multi-task incremental learning. Our method effectively addresses two key challenges: balancing task interference with knowledge sharing and maintaining modality specificity while promoting generality. Through a three-stage prompting strategy, PHP progressively refines audio-visual representations: the task-shared modality aggregating adapter enables cross-modal learning in shallow layers, the task-specific modality-shared dynamic generating adapter balances knowledge retention in middle layers, and the task-specific modality-independent prompts preserve task features in deep layers. Extensive experiments on three audio-visual tasks (AVE, AVVP, AVS and AVQA) demonstrate that our method significantly outperforms existing approaches in both preventing catastrophic forgetting and enabling effective knowledge transfer, validating the effectiveness of our work.

%% file: sec/6_ackownledgement.tex
\section{Acknowledgement}
% This work was supported by the "Pioneer" and "Leading Goose" R\&D Program of Zhejiang Province (2023C01046, 2023C01038), the National Nature Science Foundation of China (U21B2024)
This work was supported by the "Pioneer" and "Leading Goose" R\&D Program of Zhejiang Province(2023C01046, 2023C01038, 2024C01023), the National Nature Science Foundation of China (62322211, U21B2024, 62336008), Key Laboratory of Intelligent Processing Technology for Digital Music (Zhejiang Conservatory of Music), Ministry of Culture and Tourism (2023DMKLB004).

%% file: sec/X_suppl.tex
\appendix

\clearpage
\setcounter{page}{1}
\maketitlesupplementary

\section{Tasks and Datasets}
% 在这个工作中，我们将以多个视听任务来作为一个持续的数据流来模拟视听多任务增量学习。
In this work, we simulate audio-visual multi-task incremental learning by treating multiple audio-visual tasks as a continuous data stream. 
% 在这个过程中，我们采用了多个不同的视听理解类任务，包括 Audio-Visual Event localization， Audio-Visual Video Parsing 和Audio-Visual Question Answering
We employ multiple different audio-visual understanding tasks, including audio-visual event localization, audio-visual video parsing, audio-visual question answering and audio-visual segmentation.

\noindent
\textbf{Audio-Visual Event localization (AVE)}~\cite{ave} is concerned with identifying events within a video that are simultaneously visible and audible across various temporal intervals.
% We conduct an assessment of the AVE dataset [33].
% 后面接数据集概述
We conduct an assessment of the AVE dataset, which includes 4,143 videos across 28 event categories and one background category. These 10-second videos depict diverse scenarios like musical performances.

\noindent
\textbf{Audio-Visual Video Parsing (AVVP)}~\cite{avvp} aims to parse a video into temporal event sequences and categorize them as auditory, visual, or concurrently audio-visual. 
We perform experiments on the \textit{Look, Listen, and Parse} (LLP) dataset, which consists of 11,849 10-second video clips across 25 real-life categories. We utilize 10,000 clips with weak annotations for training, and 1,849 clips with detailed annotations for testing and verification.

\noindent
\textbf{Audio-Visual Question Answering (AVQA)}~\cite{music-avqa} is to answer questions by leveraging the correlations between visual objects and auditory cues. Experiments are conducted on the \textit{MUSIC-AVQA} dataset, which features over 45,000 Q\&A pairs in 9,288 videos totaling 150+ hours.

\noindent
\mo{\textbf{Audio-Visual Segmentation (AVS)}~\cite{avs} employs the Single Sound Source (S4) subset of AVSBench, comprising 4,932 videos (5 seconds each) with single sound-emitting objects. Each video aligns five 1-second audio clips and image frames, spanning 23 categories (e.g. human voice, instruments), and provides pixel-level annotations.}

\section{Experimental Setup}
\noindent
\textbf{Metrics.} 
% MEAN ACC	FINAL ACC	MEAN FGT	SINGAL ACC	MULTI ACC
% 在这个工作中，我们总体上采用准确度来表示我们的模型在三个不同的任务上的性能表现。准确度分为MEAN ACC，FINAL ACC，MEAN FGT，SINGAL ACC，MULTI ACC 五个指标。 MEAN ACC 代表了该任务在三个增量结算上的平均准确度。FINAL ACC表示了初始任务在经过增量过程后的准确度。MEAN FGT表示初始任务在增量过程后的平均遗忘率。SINGAL ACC表达了在任务单独训练时的性能。 MULTI ACC 表达了在经过了多个任务训练后，最后任务的准确度。
% 我们的目的在于通过这五个指标来有效表达出
% 1) 模型在某个任务上的具体性能。
% 2) 模型在任务增量过程中的遗忘程度。
% 3) 模型在多任务增量过程中前序任务对后续任务的有益程度。
In this work, we generally use accuracy to represent the performance of our model on three different tasks, which is divided into five metrics: $A_{mean}$, $A_{final}$, $F_{mean}$, $A_{single}$, and $A_{multi}$. $A_{mean}$ represents the average accuracy of the task over three incremental settlements. $A_{final}$ indicates the accuracy of the initial task after the incremental process. $F_{mean}$ denotes the average forgetting rate of the initial task after the incremental process. $A_{single}$ expresses the performance when the task is trained individually. $A_{multi}$ is the accuracy of the final task after training on multiple tasks.
With these five indicators, our aim is to effectively demonstrate: 1) The performance of the model on a particular task. 2) The degree of forgetting of the model during the multi-task incremental process. 3) The beneficial extent of previous tasks to subsequent tasks during the multi-task incremental process.

\noindent
\textbf{Implements details.} 
Our model builds upon the pre-trained CLIP and CLAP architectures as its backbone for handling three audio-visual downstream tasks. Specifically, we utilize a frozen CLIP-trained ViT~\cite{vit} for visual encoding and a frozen CLAP-trained HTS-AT~\cite{hstat} for audio encoding, leveraging their pre-trained parameters for robust multi-modal feature extraction. 
The prompts and adapters are strategically injected into both ViT and HTS-AT layers to facilitate audio-visual cross-modal correspondence while preserving knowledge from previous tasks. During training, we set the batch size to 3 and train each task for 10 epochs to ensure convergence.
For the optimization process, we employ the Adam optimizer with a learning rate of 3e-4 and a weight decay of 2e-4. The learning rate is scheduled with a cosine decay strategy. We conduct all experiments on a single NVIDIA 3090 GPU with 24GB memory.

\section{Normalized Penalty-aware Difference}
To better evaluate the improvement of different methods in multi-task learning scenarios, we propose a novel evaluation metric called normalized penalty-aware difference ($Diff$). This metric is designed to address two key challenges in performance evaluation: (1) the difficulty of achieving improvements upon higher baseline performance, and (2) the unfair advantage of methods with lower baseline performance showing larger absolute improvements.

The metric is defined by incorporating both normalized improvement and baseline performance through a quadratic penalty term:

\begin{equation}
   Diff = \frac{A_{multi} - A_{single}}{100 - A_{single}} \times (1 + \frac{A_{single}}{100})^2 \times 100\%
\end{equation}

where $A_{multi}$ and $A_{single}$ represent the performance of multi-task and single-task training respectively. The metric consists of three key components. First, the normalized improvement term $\frac{A_{multi} - A_{single}}{100 - A_{single}}$ considers the relative improvement potential by normalizing the performance gain against the remaining improvement headroom. Second, the quadratic baseline penalty $(1 + \frac{A_{single}}{100})^2$ introduces a penalty that grows quadratically with the baseline performance, reflecting the increasing difficulty of achieving improvements as the baseline performance gets higher. Finally, the multiplication by 100\% converts the score into a percentage form for better interpretability.

To handle boundary cases where baseline performance approaches 100\%, we introduce a small positive number $\epsilon$ (e.g., 0.001) to prevent division by zero:

\begin{equation}
   Diff = \frac{A_{multi} - A_{single}}{\max(100 - A_{single}, \epsilon)} \times (1 + \frac{A_{single}}{100})^2 \times 100\%
\end{equation}

In Table~2 in the main text, our metric provides a more nuanced evaluation of different methods' capabilities. For example, while Fine-tune shows significant performance degradation on AVE (from 57.47\% to 18.22\%) but improvement on AVVP (from 52.64\% to 59.00\%), its overall Diff score of -58.16\% indicates severe negative transfer. By comparison, EWC (-2.87\%) and PC (-3.94\%) demonstrate more stable performance but still fail to achieve positive knowledge transfer. Notably, despite PC achieving high performance on AVQA (69.85\%), its overall transfer capability remains negative. In contrast, our method is the only approach showing positive knowledge transfer (+7.79\%), with particularly strong performance on AVE (improving from 70.45\% to 72.52\%). These results clearly demonstrate that our progressive prompting approach effectively leverages knowledge from previously learned tasks to enhance performance on new tasks while maintaining robustness across both cross-task and cross-modal scenarios.

\section{More Detailed Experiments of Comparison Methods}
Here we provide more detailed experimental results to demonstrate the performance of different methods under various task orders. We evaluate six methods (Fine-tune, EWC, L2P, S-prompt, Dualprompt, and PC) across six different task ordering scenarios to track their performance through three stages of incremental learning. These comprehensive results allow us to analyze both the forgetting resistance and knowledge transfer capabilities of each method in detail.

Tab.\ref{tab:finetune_task_order}-Tab.\ref{tab:PC_task_order} present the complete performance metrics for each method. The progression of performance scores reveals distinct patterns in how different approaches handle catastrophic forgetting and knowledge transfer. When examining the performance trajectory, we observe that traditional methods like fine-tuning exhibit severe forgetting, with performance often deteriorating dramatically as new tasks are learned. For instance, in the AVE→AVVP→AVQA sequence, fine-tuning's performance on AVE drops from 56.77\% to 19.48\% after learning AVVP, and further declines to 17.79\% after learning AVQA, demonstrating significant knowledge erosion of the initial task.
In contrast, the PC method demonstrates greater resilience to catastrophic forgetting. In the same AVE→AVVP→AVQA sequence, PC maintains the performance on AVE at 54.50\% after learning all three tasks, compared to the original 69.90\%. However, PC is not immune to forgetting, as evidenced in the AVVP→AVQA→AVE sequence where the performance on AVVP drops from 45.66\% to 22.90\% after learning all tasks. Notably, the task order significantly impacts performance retention, with certain sequences like AVQA→AVVP→AVE allowing PC to maintain relatively stable performance on the first task (69.44\% from an initial 69.72\%).

Table~\ref{tab:DCNet_task_order}-~\ref{tab:S_D_M} provide an extensive analysis of different component configurations and ordering strategies across various task sequences. Table~\ref{tab:wo_TMA_TMDG} shows the detailed performance of the model without TMA-TMDG-TMI components under different task orders. Table~\ref{tab:wo_TMDG-TMI}-~\ref{tab:wo_TMA_TMDG} demonstrate the performance of models with TMDG-TMI components, TMA-TMI components, and TMA-TMDG components removed, respectively.
Table~\ref{tab:wo_TMI}-~\ref{tab:wo_TMA} focus on ablation studies of individual components. Removing TMI (Table~\ref{tab:wo_TMI}) affects task-specific representation refinement, while removing TMDG (Table~\ref{tab:wo_TMDG}) impairs cross-modal integration capabilities. Table~\ref{tab:wo_TMA} shows that without TMA, the model struggles to establish foundational audio-visual correspondences, particularly affecting performance on earlier tasks in the sequence.
Table~\ref{tab:D_M_S}-~\ref{tab:S_D_M} examine different ordering arrangements of our components across network depths: D-M-S (Table~\ref{tab:D_M_S}), M-D-S (Table~\ref{tab:M_D_S}), D-S-M (Table~\ref{tab:D_S_M}), M-S-D (Table~\ref{tab:M_S_D}), and S-D-M (Table~\ref{tab:S_D_M}) configurations, where D represents deep layers, M represents middle layers, and S represents shallow layers. The results consistently demonstrate that our proposed progressive S-M-D ordering (shallow-middle-deep) achieves optimal performance across different metrics and task sequences. This validates our key design principle: universal representations should be learned at shallow layers, task-specific cross-modal features at middle layers, and fine-grained task-modality details at deeper layers.
These comprehensive results strongly support our architectural design choices and demonstrate the effectiveness of our progressive prompting strategy for audio-visual multi-task incremental learning.

\begin{table}[tp]
  \centering
  \caption{Performance analysis of fine-tuning under different task orders. Each column within an order shows the performance of tasks as they are incrementally learned.}
  \resizebox{\linewidth}{!}{
  \begin{tabular}{l|ccc|ccc|ccc}
  \toprule[1.5pt]
  \multicolumn{10}{c}{\textbf{fine-tune}} \\
  \midrule
  \rowcolor{blue!15} Stage & \multicolumn{3}{c|}{AVE→AVVP→AVQA} & \multicolumn{3}{c|}{AVE→AVQA→AVVP} & \multicolumn{3}{c}{AVVP→AVE→AVQA} \\
  \midrule
  \rowcolor{gray!10} 1 & 56.77 & & & 58.16 & & & 51.95 & & \\
  2 & 19.48 & 50.16 & & 17.77 & 54.27 & & 46.21 & 36.79 & \\
  \rowcolor{gray!10} 3 & 17.79 & 63.01 & 54.18 & 7.69 & 53.02 & 59.71 & 63.01 & 17.79 & 54.08 \\
  \midrule
  \rowcolor{blue!15} Stage & \multicolumn{3}{c|}{AVVP→AVQA→AVE} & \multicolumn{3}{c|}{AVQA→AVE→AVVP} & \multicolumn{3}{c}{AVQA→AVVP→AVE} \\
  \midrule
  \rowcolor{gray!10} 1 & 53.33 & & & 54.16 & & & 54.22 & & \\
  2 & 63.01 & 54.21 & & 54.24 & 18.11 & & 54.00 & 58.70 & \\
  \rowcolor{gray!10} 3 & 13.58 & 53.16 & 18.21 & 52.50 & 8.98 & 58.28 & 54.47 & 10.05 & 18.23 \\
  \bottomrule[1.5pt]
  \end{tabular}}%
  \label{tab:finetune_task_order}%
  \end{table}%

\begin{table}[tp]
 \centering
 \caption{Performance analysis of EWC under different task orders. Each column within an order shows the performance of tasks as they are incrementally learned. The first task's performance is tracked across all stages to evaluate forgetting resistance, while the final task's performance demonstrates transfer capability.}
 \resizebox{\linewidth}{!}{
   \begin{tabular}{l|ccc|ccc|ccc}
   \toprule[1.5pt]
   \multicolumn{10}{c}{\textbf{EWC}} \\
   \midrule
   \rowcolor{blue!15} Stage & \multicolumn{3}{c|}{AVE→AVVP→AVQA} & \multicolumn{3}{c|}{AVE→AVQA→AVVP} & \multicolumn{3}{c}{AVVP→AVE→AVQA} \\
   \midrule
   \rowcolor{gray!10} 1     & 17.79 &       &       & 17.79 &       &       & 63.01 &       &  \\
   2     & 9.68  & 63.01 &       & 1.07  & 54.97 &       & 5.51  & 17.79 &  \\
   \rowcolor{gray!10} 3     & 1.57  & 0.92  & 54.51 & 3.18  & 54.95 & 63.01 & 0.69  & 3.93  & 54.76 \\
   \midrule
   \rowcolor{blue!15} Stage & \multicolumn{3}{c|}{AVVP→AVQA→AVE} & \multicolumn{3}{c|}{AVQA→AVE→AVVP} & \multicolumn{3}{c}{AVQA→AVVP→AVE} \\
   \midrule
   \rowcolor{gray!10} 1     & 63.01 &       &       & 54.44 &       &       & 58.95 &       &  \\
   2     & 1.61  & 52.13 &       & 54.45 & 17.79 &       & 59.12 & 62.74 &  \\
   \rowcolor{gray!10} 3     & 1.47  & 52.51 & 17.74 & 54.30 & 4.55  & 62.32 & 59.10 & 62.55 & 18.26 \\
   \bottomrule[1.5pt]
   \end{tabular}}%
 \label{tab:EWC_task_order}%
\end{table}%

\begin{table}[tp]
 \centering
 \caption{Performance analysis of L2P under different task orders. Each column within an order shows the performance of tasks as they are incrementally learned. }
 \resizebox{\linewidth}{!}{
   \begin{tabular}{l|ccc|ccc|ccc}
   \toprule[1.5pt]
   \multicolumn{10}{c}{\textbf{L2P}} \\
   \midrule
   \rowcolor{blue!15} Stage & \multicolumn{3}{c|}{AVE→AVVP→AVQA} & \multicolumn{3}{c|}{AVE→AVQA→AVVP} & \multicolumn{3}{c}{AVVP→AVE→AVQA} \\
   \midrule
   \rowcolor{gray!10} 1     & 70.32 &       &       & 69.28 &       &       & 48.69 &       &  \\
   2     & 34.83 & 48.32 &       & 69.35 & 60.03 &       & 35.47 & 65.45 &  \\
   \rowcolor{gray!10} 3     & 34.73 & 48.32 & 60.20 & 33.76 & 60.14 & 47.22 & 39.74 & 65.70 & 59.87 \\
   \midrule
   \rowcolor{blue!15} Stage & \multicolumn{3}{c|}{AVVP→AVQA→AVE} & \multicolumn{3}{c|}{AVQA→AVE→AVVP} & \multicolumn{3}{c}{AVQA→AVVP→AVE} \\
   \midrule
   \rowcolor{gray!10} 1     & 48.74 &       &       & 59.30 &       &       & 60.28 &       &  \\
   2     & 49.20 & 60.02 &       & 59.22 & 68.11 &       & 60.20 & 49.33 &  \\
   \rowcolor{gray!10} 3     & 26.71 & 59.99 & 68.41 & 59.31 & 68.21 & 42.72 & 60.28 & 38.64 & 66.69 \\
   \bottomrule[1.5pt]
   \end{tabular}}%
 \label{tab:L2P_task_order}%
\end{table}%

\begin{table}[tp]
 \centering
 \caption{Performance analysis of S-prompt under different task orders. Each column within an order shows the performance of tasks as they are incrementally learned.}
 \resizebox{\linewidth}{!}{
   \begin{tabular}{l|ccc|ccc|ccc}
   \toprule[1.5pt]
   \multicolumn{10}{c}{\textbf{S-prompt}} \\
   \midrule
   \rowcolor{blue!15} Stage & \multicolumn{3}{c|}{AVE→AVVP→AVQA} & \multicolumn{3}{c|}{AVE→AVQA→AVVP} & \multicolumn{3}{c}{AVVP→AVE→AVQA} \\
   \midrule
   \rowcolor{gray!10} 1     & 60.95 &       &       & 63.66 &       &       & 45.66 &       &  \\
   2     & 51.27 & 42.50 &       & 64.08 & 58.75 &       & 34.10 & 48.71 &  \\
   \rowcolor{gray!10} 3     & 51.09 & 42.31 & 59.85 & 54.40 & 58.76 & 42.96 & 31.99 & 17.74 & 49.29 \\
   \midrule
   \rowcolor{blue!15} Stage & \multicolumn{3}{c|}{AVVP→AVQA→AVE} & \multicolumn{3}{c|}{AVQA→AVE→AVVP} & \multicolumn{3}{c}{AVQA→AVVP→AVE} \\
   \midrule
   \rowcolor{gray!10} 1     & 51.17 &       &       & 59.30 &       &       & 58.95 &       &  \\
   2     & 50.94 & 58.04 &       & 59.22 & 59.75 &       & 59.12 & 52.64 &  \\
   \rowcolor{gray!10} 3     & 28.27 & 58.08 & 54.25 & 59.31 & 51.94 & 42.72 & 59.10 & 43.32 & 54.53 \\
   \bottomrule[1.5pt]
   \end{tabular}}%
 \label{tab:Sprompt_task_order}%
\end{table}%

\begin{table}[tp]
\centering
\caption{Performance analysis of Dualprompt under different task orders. Each column within an order shows the performance of tasks as they are incrementally learned. }
\resizebox{\linewidth}{!}{
  \begin{tabular}{l|ccc|ccc|ccc}
  \toprule[1.5pt]
  \multicolumn{10}{c}{\textbf{Dualprompt}} \\
  \midrule
  \rowcolor{blue!15} Stage & \multicolumn{3}{c|}{AVE→AVVP→AVQA} & \multicolumn{3}{c|}{AVE→AVQA→AVVP} & \multicolumn{3}{c}{AVVP→AVE→AVQA} \\
  \midrule
  \rowcolor{gray!10} 1     & 68.21 &       &       & 68.08 &       &       & 46.17 &       &  \\
  2     & 58.26 & 45.85 &       & 67.26 & 64.69 &       & 29.14 & 68.46 &  \\
  \rowcolor{gray!10} 3     & 56.84 & 42.59 & 64.57 & 59.35 & 64.82 & 46.03 & 31.07 & 67.24 & 63.77 \\
  \midrule
  \rowcolor{blue!15} Stage & \multicolumn{3}{c|}{AVVP→AVQA→AVE} & \multicolumn{3}{c|}{AVQA→AVE→AVVP} & \multicolumn{3}{c}{AVQA→AVVP→AVE} \\
  \midrule
  \rowcolor{gray!10} 1     & 45.66 &       &       & 63.48 &       &       & 63.15 &       &  \\
  2     & 46.31 & 64.06 &       & 63.54 & 68.28 &       & 63.19 & 44.75 &  \\
  \rowcolor{gray!10} 3     & 25.52 & 64.06 & 67.91 & 63.75 & 54.35 & 46.35 & 63.19 & 34.97 & 67.04 \\
  \bottomrule[1.5pt]
  \end{tabular}}%
\label{tab:Dualprompt_task_order}%
\end{table}%

\begin{table}[tp]
\centering
\caption{Performance analysis of PC under different task orders. Each column within an order shows the performance of tasks as they are incrementally learned. }
\resizebox{\linewidth}{!}{
  \begin{tabular}{l|ccc|ccc|ccc}
  \toprule[1.5pt]
  \multicolumn{10}{c}{\textbf{PC}} \\
  \midrule
  \rowcolor{blue!15} Stage & \multicolumn{3}{c|}{AVE→AVVP→AVQA} & \multicolumn{3}{c|}{AVE→AVQA→AVVP} & \multicolumn{3}{c}{AVVP→AVE→AVQA} \\
  \midrule
  \rowcolor{gray!10} 1     & 69.90 &       &       & 69.43 &       &       & 44.42 &       &  \\
  2     & 55.50 & 45.89 &       & 68.38 & 69.24 &       & 36.90 & 66.57 &  \\
  \rowcolor{gray!10} 3     & 54.50 & 45.53 & 69.86 & 54.38 & 69.04 & 44.56 & 39.06 & 66.97 & 69.84 \\
  \midrule
  \rowcolor{blue!15} Stage & \multicolumn{3}{c|}{AVVP→AVQA→AVE} & \multicolumn{3}{c|}{AVQA→AVE→AVVP} & \multicolumn{3}{c}{AVQA→AVVP→AVE} \\
  \midrule
  \rowcolor{gray!10} 1     & 45.66 &       &       & 69.35 &       &       & 69.72 &       &  \\
  2     & 45.48 & 69.67 &       & 69.51 & 71.00 &       & 69.78 & 44.47 &  \\
  \rowcolor{gray!10} 3     & 22.90 & 69.62 & 68.86 & 69.47 & 56.27 & 44.65 & 69.44 & 30.61 & 67.21 \\
  \bottomrule[1.5pt]
  \end{tabular}}%
\label{tab:PC_task_order}%
\end{table}%

\begin{table}[tp]
  \centering
  \caption{Performance analysis of DCNet under different task orders. Each column within an order shows the performance of tasks as they are incrementally learned. }
  \resizebox{\linewidth}{!}{
    \begin{tabular}{l|ccc|ccc|ccc}
    \toprule[1.5pt]
    \multicolumn{10}{c}{\textbf{DCNet}} \\
    \midrule
    \rowcolor{blue!15} Stage & \multicolumn{3}{c|}{AVE→AVVP→AVQA} & \multicolumn{3}{c|}{AVE→AVQA→AVVP} & \multicolumn{3}{c}{AVVP→AVE→AVQA} \\
    \midrule
    \rowcolor{gray!10} 1     & 58.13 &       &       & 60.52 &       &       & 54.34 &       &  \\
    2     & 20.62 & 48.83 &       & 17.79 & 54.25 &       & 46.73 & 38.06 &  \\
    \rowcolor{gray!10} 3     & 17.79 & 63.01 & 53.93 & 7.71 & 53.57 & 59.48 & 63.01 & 17.79 & 54.01 \\
    \midrule
    \rowcolor{blue!15} Stage & \multicolumn{3}{c|}{AVVP→AVQA→AVE} & \multicolumn{3}{c|}{AVQA→AVE→AVVP} & \multicolumn{3}{c}{AVQA→AVVP→AVE} \\
    \midrule
    \rowcolor{gray!10} 1     & 50.07 &       &       & 54.23 &       &       & 54.13 &       &  \\
    2     & 63.01 & 54.37 &       & 54.47 & 17.71 &       & 53.14 & 59.57 &  \\
    \rowcolor{gray!10} 3     & 15.33 & 50.38 & 18.28 & 54.20 & 5.92 & 55.71 & 53.52 & 7.71 & 19.83 \\
    \bottomrule[1.5pt]
    \end{tabular}}%
  \label{tab:DCNet_task_order}%
 \end{table}%

 \begin{table}[tp]
  \centering
  \caption{Ablation study: Detailed Performance of model without TMA-TMDG-TMI components under different task orders. Each column within an order shows the performance of tasks as they are incrementally learned.}
  \resizebox{\linewidth}{!}{
    \begin{tabular}{l|ccc|ccc|ccc}
    \toprule[1.5pt]
    \multicolumn{10}{c}{\textbf{w/o TMA-TMDG-TMI}} \\
    \midrule
    \rowcolor{blue!15} Stage & \multicolumn{3}{c|}{AVE→AVVP→AVQA} & \multicolumn{3}{c|}{AVE→AVQA→AVVP} & \multicolumn{3}{c}{AVVP→AVE→AVQA} \\
    \midrule
    \rowcolor{gray!10} 1     & 70.67 &       &       & 70.87 &       &       & 48.14 &       &  \\
    2     & 57.46 & 45.80 &       & 70.62 & 69.39 &       & 32.35 & 69.03 &  \\
    \rowcolor{gray!10} 3     & 58.96 & 45.76 & 69.82 & 57.56 & 69.61 & 46.49 & 34.33 & 69.25 & 70.17 \\
    \midrule
    \rowcolor{blue!15} Stage & \multicolumn{3}{c|}{AVVP→AVQA→AVE} & \multicolumn{3}{c|}{AVQA→AVE→AVVP} & \multicolumn{3}{c}{AVQA→AVVP→AVE} \\
    \midrule
    \rowcolor{gray!10} 1     & 46.90 &       &       & 69.28 &       &       & 69.85 &       &  \\
    2     & 47.04 & 69.94 &       & 69.60 & 69.78 &       & 69.48 & 47.64 &  \\
    \rowcolor{gray!10} 3     & 35.34 & 69.98 & 68.06 & 69.36 & 57.11 & 45.20 & 69.76 & 30.29 & 70.80 \\
    \bottomrule[1.5pt]
    \end{tabular}}%
  \label{tab:wo_TMA_TMDG_TMI}%
 \end{table}%

 \begin{table}[tp]
  \centering
  \caption{Ablation study: Detailed Performance without TMDG-TMI components under different task orders.}
  \resizebox{\linewidth}{!}{
    \begin{tabular}{l|ccc|ccc|ccc}
    \toprule[1.5pt]
    \multicolumn{10}{c}{\textbf{w/o TMDG-TMI}} \\
    \midrule
    \rowcolor{blue!15} Stage & \multicolumn{3}{c|}{AVE→AVVP→AVQA} & \multicolumn{3}{c|}{AVE→AVQA→AVVP} & \multicolumn{3}{c}{AVVP→AVE→AVQA} \\
    \midrule
    \rowcolor{gray!10} 1     & 68.91 &       &       & 69.45 &       &       & 45.53 &       &  \\
    2     & 60.25 & 46.49 &       & 68.88 & 70.06 &       & 29.88 & 69.70 &  \\
    \rowcolor{gray!10} 3     & 62.79 & 51.95 & 70.17 & 59.65 & 69.72 & 48.55 & 35.57 & 63.48 & 70.27 \\
    \midrule
    \rowcolor{blue!15} Stage & \multicolumn{3}{c|}{AVVP→AVQA→AVE} & \multicolumn{3}{c|}{AVQA→AVE→AVVP} & \multicolumn{3}{c}{AVQA→AVVP→AVE} \\
    \midrule
    \rowcolor{gray!10} 1     & 46.81 &       &       & 69.72 &       &       & 70.08 &       &  \\
    2     & 49.34 & 70.71 &       & 69.27 & 71.49 &       & 69.53 & 48.97 &  \\
    \rowcolor{gray!10} 3     & 34.19 & 70.55 & 70.99 & 69.10 & 59.65 & 46.77 & 69.10 & 34.37 & 70.82 \\
    \bottomrule[1.5pt]
    \end{tabular}}
    \label{tab:wo_TMDG-TMI}%
  \end{table}%

\begin{table}[tp]
  \centering
  \caption{Ablation study: Detailed Performance without TMA-TMI components under different task orders. }
  \resizebox{\linewidth}{!}{
    \begin{tabular}{l|ccc|ccc|ccc}
    \toprule[1.5pt]
    \multicolumn{10}{c}{\textbf{w/o TMA-TMI}} \\
    \midrule
    \rowcolor{blue!15} Stage & \multicolumn{3}{c|}{AVE→AVVP→AVQA} & \multicolumn{3}{c|}{AVE→AVQA→AVVP} & \multicolumn{3}{c}{AVVP→AVE→AVQA} \\
    \midrule
    \rowcolor{gray!10} 1     & 70.03 &       &       & 71.19 &       &       & 46.95 &       &  \\
    2     & 60.32 & 47.45 &       & 70.97 & 69.21 &       & 43.97 & 70.20 &  \\
    \rowcolor{gray!10} 3     & 61.77 & 47.13 & 70.07 & 59.15 & 69.30 & 47.32 & 39.70 & 69.88 & 69.49 \\
    \midrule
    \rowcolor{blue!15} Stage & \multicolumn{3}{c|}{AVVP→AVQA→AVE} & \multicolumn{3}{c|}{AVQA→AVE→AVVP} & \multicolumn{3}{c}{AVQA→AVVP→AVE} \\
    \midrule
    \rowcolor{gray!10} 1     & 45.76 &       &       & 69.69 &       &       & 69.57 &       &  \\
    2     & 45.53 & 69.47 &       & 69.36 & 69.13 &       & 69.73 & 48.28 &  \\
    \rowcolor{gray!10} 3     & 32.58 & 69.65 & 69.58 & 69.65 & 59.58 & 46.58 & 69.64 & 30.70 & 69.83 \\
    \bottomrule[1.5pt]
    \end{tabular}}%
  \label{tab:wo_TMA_TMI}%
  \end{table}%

  \begin{table}[tp]
  \centering
  \caption{Ablation study: Detailed performance without TMA-TMDG components under different task orders.}
  \resizebox{\linewidth}{!}{
    \begin{tabular}{l|ccc|ccc|ccc}
    \toprule[1.5pt]
    \multicolumn{10}{c}{\textbf{w/o TMA-TMDG}} \\
    \midrule
    \rowcolor{blue!15} Stage & \multicolumn{3}{c|}{AVE→AVVP→AVQA} & \multicolumn{3}{c|}{AVE→AVQA→AVVP} & \multicolumn{3}{c}{AVVP→AVE→AVQA} \\
    \midrule
    \rowcolor{gray!10} 1     & 71.79 &       &       & 70.65 &       &       & 50.12 &       &  \\
    2     & 62.31 & 49.01 &       & 70.57 & 69.55 &       & 39.01 & 71.42 &  \\
    \rowcolor{gray!10} 3     & 63.06 & 48.55 & 68.76 & 55.30 & 69.69 & 49.29 & 41.53 & 71.12 & 69.20 \\
    \midrule
    \rowcolor{blue!15} Stage & \multicolumn{3}{c|}{AVVP→AVQA→AVE} & \multicolumn{3}{c|}{AVQA→AVE→AVVP} & \multicolumn{3}{c}{AVQA→AVVP→AVE} \\
    \midrule
    \rowcolor{gray!10} 1     & 48.88 &       &       & 69.95 &       &       & 69.08 &       &  \\
    2     & 48.60 & 68.99 &       & 69.53 & 72.26 &       & 69.15 & 50.02 &  \\
    \rowcolor{gray!10} 3     & 34.65 & 68.90 & 71.37 & 69.69 & 62.36 & 48.23 & 70.81 & 36.99 & 73.21 \\
    \bottomrule[1.5pt]
    \end{tabular}}%
  \label{tab:wo_TMA_TMDG}%
  \end{table}%

     \begin{table}[tp]
      \centering
      \caption{Ablation study: Detailed Performance without TMI component under different task orders. }
      \resizebox{\linewidth}{!}{
        \begin{tabular}{l|ccc|ccc|ccc}
        \toprule[1.5pt]
        \multicolumn{10}{c}{\textbf{w/o TMI}} \\
        \midrule
        \rowcolor{blue!15} Stage & \multicolumn{3}{c|}{AVE→AVVP→AVQA} & \multicolumn{3}{c|}{AVE→AVQA→AVVP} & \multicolumn{3}{c}{AVVP→AVE→AVQA} \\
        \midrule
        \rowcolor{gray!10} 1     & 69.65 &       &       & 68.09 &       &       & 48.88 &       &  \\
        2     & 58.16 & 47.50 &       & 68.33 & 69.68 &       & 45.07 & 71.37 &  \\
        \rowcolor{gray!10} 3     & 59.78 & 48.92 & 70.47 & 57.81 & 69.54 & 48.80 & 45.98 & 66.34 & 70.09 \\
        \midrule
        \rowcolor{blue!15} Stage & \multicolumn{3}{c|}{AVVP→AVQA→AVE} & \multicolumn{3}{c|}{AVQA→AVE→AVVP} & \multicolumn{3}{c}{AVQA→AVVP→AVE} \\
        \midrule
        \rowcolor{gray!10} 1     & 47.87 &       &       & 70.10 &       &       & 70.05 &       &  \\
        2     & 50.12 & 70.02 &       & 69.44 & 71.29 &       & 69.81 & 50.21 &  \\
        \rowcolor{gray!10} 3     & 33.73 & 69.92 & 71.22 & 69.55 & 57.44 & 47.18 & 69.23 & 36.90 & 70.95 \\
        \bottomrule[1.5pt]
        \end{tabular}}%
      \label{tab:wo_TMI}%
     \end{table}%

     \begin{table}[tp]
      \centering
      \caption{Ablation study: Detailed Performance without TMDG component under different task orders.}
      \resizebox{\linewidth}{!}{
        \begin{tabular}{l|ccc|ccc|ccc}
        \toprule[1.5pt]
        \multicolumn{10}{c}{\textbf{w/o TMDG}} \\
        \midrule
        \rowcolor{blue!15} Stage & \multicolumn{3}{c|}{AVE→AVVP→AVQA} & \multicolumn{3}{c|}{AVE→AVQA→AVVP} & \multicolumn{3}{c}{AVVP→AVE→AVQA} \\
        \midrule
        \rowcolor{gray!10} 1     & 69.25 &       &       & 69.85 &       &       & 49.52 &       &  \\
        2     & 58.38 & 47.73 &       & 68.33 & 70.09 &       & 40.71 & 70.52 &  \\
        \rowcolor{gray!10} 3     & 58.33 & 49.24 & 70.94 & 56.19 & 69.93 & 50.53 & 42.59 & 63.61 & 70.68 \\
        \midrule
        \rowcolor{blue!15} Stage & \multicolumn{3}{c|}{AVVP→AVQA→AVE} & \multicolumn{3}{c|}{AVQA→AVE→AVVP} & \multicolumn{3}{c}{AVQA→AVVP→AVE} \\
        \midrule
        \rowcolor{gray!10} 1     & 47.41 &       &       & 70.17 &       &       & 70.16 &       &  \\
        2     & 46.54 & 71.01 &       & 69.41 & 73.04 &       & 68.83 & 48.88 &  \\
        \rowcolor{gray!10} 3     & 35.66 & 70.39 & 72.96 & 69.49 & 62.24 & 48.97 & 68.93 & 35.20 & 71.47 \\
        \bottomrule[1.5pt]
        \end{tabular}}%
      \label{tab:wo_TMDG}%
     \end{table}%

     \begin{table}[tp]
      \centering
      \caption{Ablation study: Detailed Performance without TMA component under different task orders. }
      \resizebox{\linewidth}{!}{
        \begin{tabular}{l|ccc|ccc|ccc}
        \toprule[1.5pt]
        \multicolumn{10}{c}{\textbf{w/o TMA}} \\
        \midrule
        \rowcolor{blue!15} Stage & \multicolumn{3}{c|}{AVE→AVVP→AVQA} & \multicolumn{3}{c|}{AVE→AVQA→AVVP} & \multicolumn{3}{c}{AVVP→AVE→AVQA} \\
        \midrule
        \rowcolor{gray!10} 1     & 70.47 &       &       & 70.17 &       &       & 49.38 &       &  \\
        2     & 63.31 & 49.11 &       & 70.37 & 70.01 &       & 31.02 & 69.75 &  \\
        \rowcolor{gray!10} 3     & 61.00 & 48.10 & 69.73 & 60.42 & 69.81 & 49.01 & 37.08 & 69.58 & 69.95 \\
        \midrule
        \rowcolor{blue!15} Stage & \multicolumn{3}{c|}{AVVP→AVQA→AVE} & \multicolumn{3}{c|}{AVQA→AVE→AVVP} & \multicolumn{3}{c}{AVQA→AVVP→AVE} \\
        \midrule
        \rowcolor{gray!10} 1     & 50.12 &       &       & 70.13 &       &       & 69.56 &       &  \\
        2     & 50.02 & 69.77 &       & 70.05 & 71.69 &       & 69.43 & 49.75 &  \\
        \rowcolor{gray!10} 3     & 34.88 & 69.66 & 71.19 & 69.84 & 60.95 & 48.69 & 69.81 & 30.93 & 70.67 \\
        \bottomrule[1.5pt]
        \end{tabular}}%
      \label{tab:wo_TMA}%
     \end{table}%

     \begin{table}[tp]
      \centering
      \caption{Ablation study: Detailed Performance with D-M-S component ordering under different task orders. }
      \resizebox{\linewidth}{!}{
        \begin{tabular}{l|ccc|ccc|ccc}
        \toprule[1.5pt]
        \multicolumn{10}{c}{\textbf{D-M-S}} \\
        \midrule
        \rowcolor{blue!15} Stage & \multicolumn{3}{c|}{AVE→AVVP→AVQA} & \multicolumn{3}{c|}{AVE→AVQA→AVVP} & \multicolumn{3}{c}{AVVP→AVE→AVQA} \\
        \midrule
        \rowcolor{gray!10} 1     & 51.12 &       &       & 51.82 &       &       & 52.69 &       &  \\
        2     & 31.57 & 51.17 &       & 18.21 & 63.36 &       & 35.89 & 50.65 &  \\
        \rowcolor{gray!10} 3     & 14.42 & 59.29 & 63.97 & 26.19 & 56.71 & 50.76 & 43.23 & 18.31 & 63.94 \\
        \midrule
        \rowcolor{blue!15} Stage & \multicolumn{3}{c|}{AVVP→AVQA→AVE} & \multicolumn{3}{c|}{AVQA→AVE→AVVP} & \multicolumn{3}{c}{AVQA→AVVP→AVE} \\
        \midrule
        \rowcolor{gray!10} 1     & 51.72 &       &       & 63.52 &       &       & 63.43 &       &  \\
        2     & 63.10 & 63.91 &       & 58.20 & 50.03 &       & 58.42 & 53.60 &  \\
        \rowcolor{gray!10} 3     & 46.90 & 60.08 & 51.64 & 56.67 & 31.49 & 50.71 & 59.41 & 35.75 & 51.57 \\
        \bottomrule[1.5pt]
        \end{tabular}}%
      \label{tab:D_M_S}%
     \end{table}%

     \begin{table}[tp]
      \centering
      \caption{Ablation study: Detailed Performance with M-D-S component ordering under different task orders. }
      \resizebox{\linewidth}{!}{
        \begin{tabular}{l|ccc|ccc|ccc}
        \toprule[1.5pt]
        \multicolumn{10}{c}{\textbf{M-D-S}} \\
        \midrule
        \rowcolor{blue!15} Stage & \multicolumn{3}{c|}{AVE→AVVP→AVQA} & \multicolumn{3}{c|}{AVE→AVQA→AVVP} & \multicolumn{3}{c}{AVVP→AVE→AVQA} \\
        \midrule
        \rowcolor{gray!10} 1     & 55.10 &       &       & 54.81 &       &       & 48.60 &       &  \\
        2     & 34.55 & 52.32 &       & 18.07 & 63.94 &       & 31.99 & 52.84 &  \\
        \rowcolor{gray!10} 3     & 12.51 & 63.01 & 64.52 & 29.58 & 58.89 & 55.21 & 62.92 & 19.23 & 64.51 \\
        \midrule
        \rowcolor{blue!15} Stage & \multicolumn{3}{c|}{AVVP→AVQA→AVE} & \multicolumn{3}{c|}{AVQA→AVE→AVVP} & \multicolumn{3}{c}{AVQA→AVVP→AVE} \\
        \midrule
        \rowcolor{gray!10} 1     & 50.48 &       &       & 63.22 &       &       & 63.51 &       &  \\
        2     & 63.39 & 63.97 &       & 58.55 & 52.81 &       & 61.02 & 55.12 &  \\
        \rowcolor{gray!10} 3     & 37.31 & 60.97 & 51.54 & 61.25 & 36.42 & 51.63 & 61.46 & 40.29 & 55.05 \\
        \bottomrule[1.5pt]
        \end{tabular}}%
      \label{tab:M_D_S}%
     \end{table}%

     \begin{table}[tp]
      \centering
      \caption{Ablation study: Detailed Performance with D-S-M component ordering under different task orders. }
      \resizebox{\linewidth}{!}{
        \begin{tabular}{l|ccc|ccc|ccc}
        \toprule[1.5pt]
        \multicolumn{10}{c}{\textbf{D-S-M}} \\
        \midrule
        \rowcolor{blue!15} Stage & \multicolumn{3}{c|}{AVE→AVVP→AVQA} & \multicolumn{3}{c|}{AVE→AVQA→AVVP} & \multicolumn{3}{c}{AVVP→AVE→AVQA} \\
        \midrule
        \rowcolor{gray!10} 1     & 56.72 &       &       & 57.44 &       &       & 51.26 &       &  \\
        2     & 37.76 & 50.90 &       & 16.00 & 62.67 &       & 44.21 & 56.84 &  \\
        \rowcolor{gray!10} 3     & 18.98 & 63.01 & 54.44 & 30.67 & 57.64 & 53.69 & 51.10 & 19.35 & 63.31 \\
        \midrule
        \rowcolor{blue!15} Stage & \multicolumn{3}{c|}{AVVP→AVQA→AVE} & \multicolumn{3}{c|}{AVQA→AVE→AVVP} & \multicolumn{3}{c}{AVQA→AVVP→AVE} \\
        \midrule
        \rowcolor{gray!10} 1     & 50.80 &       &       & 54.34 &       &       & 54.44 &       &  \\
        2     & 62.23 & 64.80 &       & 54.34 & 50.03 &       & 54.44 & 52.96 &  \\
        \rowcolor{gray!10} 3     & 40.16 & 61.06 & 54.18 & 54.34 & 31.49 & 50.94 & 54.44 & 38.18 & 55.72 \\
        \bottomrule[1.5pt]
        \end{tabular}}%
      \label{tab:D_S_M}%
     \end{table}%

     \begin{table}[tp]
      \centering
      \caption{Ablation study: Detailed Performance with M-S-D component ordering under different task orders. }
      \resizebox{\linewidth}{!}{
        \begin{tabular}{l|ccc|ccc|ccc}
        \toprule[1.5pt]
        \multicolumn{10}{c}{\textbf{M-S-D}} \\
        \midrule
        \rowcolor{blue!15} Stage & \multicolumn{3}{c|}{AVE→AVVP→AVQA} & \multicolumn{3}{c|}{AVE→AVQA→AVVP} & \multicolumn{3}{c}{AVVP→AVE→AVQA} \\
        \midrule
        \rowcolor{gray!10} 1     & 55.47 &       &       & 56.39 &       &       & 52.78 &       &  \\
        2     & 39.29 & 52.13 &       & 17.74 & 62.55 &       & 40.80 & 56.89 &  \\
        \rowcolor{gray!10} 3     & 15.60 & 60.53 & 63.01 & 22.91 & 56.70 & 54.75 & 61.96 & 17.71 & 62.56 \\
        \midrule
        \rowcolor{blue!15} Stage & \multicolumn{3}{c|}{AVVP→AVQA→AVE} & \multicolumn{3}{c|}{AVQA→AVE→AVVP} & \multicolumn{3}{c}{AVQA→AVVP→AVE} \\
        \midrule
        \rowcolor{gray!10} 1     & 55.12 &       &       & 62.83 &       &       & 62.70 &       &  \\
        2     & 61.40 & 62.94 &       & 62.55 & 46.07 &       & 61.10 & 53.51 &  \\
        \rowcolor{gray!10} 3     & 36.53 & 60.05 & 53.31 & 58.27 & 32.99 & 54.89 & 61.12 & 36.12 & 51.07 \\
        \bottomrule[1.5pt]
        \end{tabular}}%
      \label{tab:M_S_D}%
     \end{table}%

     \begin{table}[tp]
      \centering
      \caption{Ablation study: Detailed Performance with S-D-M component ordering under different task orders. }
      \resizebox{\linewidth}{!}{
        \begin{tabular}{l|ccc|ccc|ccc}
        \toprule[1.5pt]
        \multicolumn{10}{c}{\textbf{S-D-M}} \\
        \midrule
        \rowcolor{blue!15} Stage & \multicolumn{3}{c|}{AVE→AVVP→AVQA} & \multicolumn{3}{c|}{AVE→AVQA→AVVP} & \multicolumn{3}{c}{AVVP→AVE→AVQA} \\
        \midrule
        \rowcolor{gray!10} 1     & 46.77 &       &       & 48.31 &       &       & 53.01 &       &  \\
        2     & 35.70 & 53.74 &       & 31.52 & 62.91 &       & 32.31 & 50.32 &  \\
        \rowcolor{gray!10} 3     & 28.06 & 53.83 & 62.70 & 28.88 & 60.49 & 54.20 & 51.77 & 30.70 & 62.95 \\
        \midrule
        \rowcolor{blue!15} Stage & \multicolumn{3}{c|}{AVVP→AVQA→AVE} & \multicolumn{3}{c|}{AVQA→AVE→AVVP} & \multicolumn{3}{c}{AVQA→AVVP→AVE} \\
        \midrule
        \rowcolor{gray!10} 1     & 52.50 &       &       & 63.60 &       &       & 62.65 &       &  \\
        2     & 56.13 & 61.54 &       & 59.11 & 45.15 &       & 60.05 & 52.13 &  \\
        \rowcolor{gray!10} 3     & 28.68 & 60.64 & 48.26 & 61.20 & 35.77 & 52.59 & 59.89 & 49.47 & 49.20 \\
        \bottomrule[1.5pt]
        \end{tabular}}%
      \label{tab:S_D_M}%
     \end{table}%